\pdfoutput=1
\documentclass{article}

\PassOptionsToPackage{square,numbers,sort&compress}{natbib}
\usepackage[preprint]{neurips_2026}

\usepackage[utf8]{inputenc}
\usepackage[T1]{fontenc}
\usepackage{hyperref}
\usepackage{url}
\usepackage{booktabs}
\usepackage{amsfonts}
\usepackage{amsmath}
\usepackage{nicefrac}
\usepackage{microtype}
\usepackage{xcolor}
\usepackage{graphicx}
\usepackage{placeins}

\newcommand{\std}[1]{\,{\color{gray}$\pm$#1}}
\newcommand{\eqv}{$^*$}
\newcommand{\pre}{$^\dagger$}

\title{How Far Can Foundation Models Be Distilled for Efficient Jet Tagging?}

\author{%
  Jay Agarwal\\
  BITS Pilani, KEK
  \And
  Siddharth Khare\\
  BITS Pilani
  \And
  Dhruv Kumar\\
  BITS Pilani
}

\begin{document}

\maketitle

\begin{abstract}
Jet tagging is one of the most important applications of machine learning at the Large Hadron Collider.
Recent models build in more physics knowledge or rely on large-scale pretraining, but the experiments place strict limits on their inference cost.
Knowledge distillation can transfer their performance to small students that are cheap to run.
We ask which properties of a teacher help the student, how small a student can get and how close it can come to the state of the art.
On a jet tagging benchmark, we distil six teachers that differ in architecture, symmetry and pretraining, including two foundation models, into five small students.
Every teacher improves every student, and the gain depends mainly on the strength of the teacher.
We then shrink our two strongest distilled students, Transformer and L-GATr-slim, in width and in depth.
With 20 thousand parameters, a distilled L-GATr-slim outperforms every published jet tagger trained from scratch at a fraction of the inference cost of every teacher.
Even with two thousand parameters, it matches ParT, a model with a thousand times as many parameters.
We also study what the students learn from their teachers.
Our benchmark of inference cost shows that students of the same size can differ by almost an order of magnitude in speed.
We release the code and the weights of all trained models.
\end{abstract}

\section{Introduction}
\label{sec:intro}

Modern machine learning has transformed data analysis at the Large Hadron Collider (LHC), and jet tagging is among its most established applications~\citep{kasieczka2019toptagging}.
A jet is a collimated spray of particles recorded by the detector.
The task is to identify the particle that a jet originated from, using the momenta of up to a few hundred measured constituents.
In the past few years, jet tagging has seen many new model designs: some adapt modern architectures with physics knowledge~\citep{qu2020particlenet,qu2022part,wu2024mipart,esmail2025iaformer}, and some build the symmetries of particle physics data into the model itself~\citep{gong2022lorentznet,bogatskiy2022pelican,spinner2024lgatr,spinner2025lloca}.
Recently, foundation models have been pretrained on large datasets, and fine-tuning them outperforms training the same model from scratch~\citep{qu2022part,bhimji2025omnilearned,hallin2025foundation}.

However, these developments have made taggers more expensive to run.
While training a tagger takes at most a few days, the experiments at the LHC place strict requirements on its inference cost~\citep{favaro2026virtues}.
Interactions between every pair of particles, equivariant layers and large pretrained models all make a tagger more expensive to evaluate, and its parameter count does not show by how much.

Knowledge distillation, or simply distillation, has become a standard technique in machine learning to reduce the size and computational cost of a model while closely matching its performance~\citep{hinton2015distilling,gou2021survey}.
A large teacher model guides the training of a small student model.
In jet tagging, the teacher is trained once, the student learns to match the teacher's predictions and only the student is deployed.

Several questions about distillation in jet tagging remain open.
Which properties of a teacher help the student: its strength, its symmetry, its pretraining or its architecture?
How does the benefit depend on the size of the student, and how small can the student get before distillation stops helping?
How close can a distilled student come to the state of the art?
Previous studies in particle physics use one or two teachers and simple students such as Deep Sets or an MLP~\citep{liu2023kd,benyas2026accelerators,gao2026jetcord}.
Because their teachers differ in strength as well as in the property of interest, they cannot tell the two effects apart.
Their students also remain well below the strongest taggers, which leaves open whether distillation costs too much performance to be useful.

In this paper we answer these questions with a systematic and controlled study on the top-tagging benchmark~\citep{kasieczka2019toptagging}.
In Section~\ref{sec:matrix} we distil six teachers into five students with about 200 thousand parameters each, about a tenth of the size of most teachers.
The teachers span a range of strength, differ in architecture, symmetry and pretraining, and include state-of-the-art models.
The students include a plain Transformer and smaller versions of the teachers.
We keep the distillation setting the same for every pair, so that the comparison is fair.
In Section~\ref{sec:scaling} we scale our two strongest distilled students, which are also among the cheapest to run, Transformer and the symmetry-aware L-GATr-slim~\citep{petitjean2025economical}, in width and in depth from about one thousand to 400 thousand parameters.
In Section~\ref{sec:learn} we study what the student learns from its teacher: how closely it reproduces the teacher's predictions and which of the teacher's mistakes it shares, down to single jets.
In Section~\ref{sec:inference} we benchmark the inference cost of every model.
We report the number of operations per jet, the throughput and peak memory on a GPU and the latency for a single jet on one CPU core.

We find that distillation improves each of our five students with every teacher, and that the benefit follows mainly the strength of the teacher, whatever its architecture or pretraining.
A teacher trained from scratch that is as strong as a pretrained one gives students that perform about as well.
Distillation also works well for small students, which can outperform much larger state-of-the-art teachers.
With about 20 thousand parameters, a distilled L-GATr-slim outperforms every published tagger trained from scratch~\citep{petitjean2025economical}, and only pretrained taggers do better.
It does so at a fraction of the inference cost of every teacher.
With two thousand parameters, a distilled L-GATr-slim still matches ParT, which has 2.1 million parameters.
Students copy their teachers closely, down to mistakes that only their own teacher makes.
Students of the same size can differ by almost an order of magnitude in speed, mainly because some of them compute features for every pair of particles.
Our code and the weights of all trained models are available at \url{https://github.com/JayAgarwal25/distilling-jet-taggers}, so that every result can be reproduced and other teachers and students can be distilled in the same way.

\section{Related work}
\label{sec:related}

\paragraph{Jet tagging}
Jet tagging has moved from calorimeter images~\citep{deoliveira2016jetimages,kasieczka2017deeptop} to sets of particles, processed by Deep Sets~\citep{komiske2019efn}, graph networks~\citep{qu2020particlenet} and transformers~\citep{mikuni2021pct,qu2022part}.
The Particle Transformer (ParT) adds pair features, quantities computed for every pair of particles such as their angular distance, to the attention scores~\citep{qu2022part}, and the More-Interaction Particle Transformer (MIParT)~\citep{wu2024mipart} and IAFormer~\citep{esmail2025iaformer} process these pair interactions in additional attention layers.
A second line of work builds the Lorentz symmetry, the invariance of physics under rotations and boosts between reference frames, into the network~\citep{butter2018lola,bogatskiy2020lgn}.
LorentzNet~\citep{gong2022lorentznet} and PELICAN~\citep{bogatskiy2022pelican} are graph networks on Lorentz invariants.
The Lorentz Geometric Algebra Transformer (L-GATr)~\citep{spinner2024lgatr} and the Lorentz Local Canonicalization (LLoCa) transformer~\citep{spinner2025lloca} are Lorentz-equivariant transformers, built on a geometric-algebra representation and on local reference frames per particle.
L-GATr-slim keeps only the scalar and vector channels of L-GATr and matches its performance at a lower cost~\citep{petitjean2025economical}.
Our teachers and students include both pair interactions and Lorentz equivariance.

\paragraph{Foundation models and inference cost}
Pretraining on large jet datasets improves tagging after fine-tuning~\citep{hallin2025foundation}, for ParT pretrained on JetClass, a simulated dataset of 100 million jets in ten classes~\citep{qu2022part}, and for dedicated foundation models such as OmniLearn~\citep{mikuni2024omnilearn}, OmniJet-$\alpha$~\citep{birk2024omnijet}, masked particle modelling~\citep{golling2024mpm} and OmniLearned~\citep{bhimji2025omnilearned}.
These studies report performance and data efficiency, while the cost of running such models has received less attention.
Equivariant and standard transformers have been compared at fixed inference cost on GPUs and CPUs~\citep{favaro2026virtues}.
L-GATr-slim and the LLoCa transformer have been shrunk to one thousand parameters and quantized~\citep{petitjean2025economical}, and ParT and MIParT keep their performance with less than half of their parameters~\citep{cheng2026redundancy}.

\paragraph{Knowledge distillation}
Distillation trains a small student to reproduce a large teacher~\citep{bucilua2006compression,ba2014deep,hinton2015distilling,gou2021survey}.
Besides the true class of each training example, its hard label, the student learns from the predicted probabilities of the teacher, its soft labels.
The soft labels tell the student how confident the teacher is on each example, which the hard labels do not, and a temperature softens them so that this information carries more weight~\citep{hinton2015distilling}.
In binary tagging the soft label is the probability the teacher assigns to the signal class, so it ranks jets by how signal-like the teacher finds them.
A stronger teacher does not always give a better student. When the teacher is much larger than the student, the student can fail to follow it~\citep{cho2019efficacy,mirzadeh2020ta}, and scaling laws for distillation show the same effect~\citep{busbridge2025distillation}.
Our smallest students have about a thousandth of the parameters of their teachers, so we can look for this gap in jet tagging.
Distillation can also pass the inductive bias of one architecture to a student of another, for example from convolutional networks to transformers~\citep{abnar2020inductive,touvron2021deit}.
A student also reproduces its teacher only in part~\citep{stanton2021does} and inherits some of its errors~\citep{ojha2023knowledge}, which we study jet by jet in Section~\ref{sec:learn}.

\paragraph{Distillation in particle physics}
Distilling LorentzNet into smaller taggers gives students that perform better and are more robust to Lorentz boosts~\citep{liu2023kd}.
DistillNet compresses a graph network into a per-particle network~\citep{bal2024distillnet}, and a normalizing flow for anomaly detection has been distilled~\citep{tahseen2026flows}.
The closest work distils OmniLearned-L, a foundation model with 423 million parameters, into small students~\citep{benyas2026accelerators}.
Its strongest student, which we call Deep Sets + GNN, has 17 thousand parameters and adds one graph neural network layer to a Deep Sets network.
That study distils one student architecture from two teachers and attributes the better student to pretraining, although its pretrained teacher is also the stronger one.
With six teachers and five students, we find that strength accounts for most of this effect (Section~\ref{sec:matrix}).
Our distilled students also outperform Deep Sets + GNN at a fraction of its operations per jet, even with a weaker teacher (Section~\ref{sec:inference}).

\section{Experimental setup}
\label{sec:setup}

\paragraph{Data}
At the LHC, colliding protons produce quarks and gluons.
These are never observed on their own. Each turns into tens to hundreds of particles, such as pions, kaons and photons, that fly in nearly the same direction.
The detector records the energy and direction of each of these particles, and a clustering algorithm groups the particles that belong together into a jet.
The top quark is different, because it is heavy enough to decay before it forms a jet, into a bottom quark and a $W$ boson that itself decays into two quarks.
Its jet therefore usually contains three dense clusters of particles, while the jet of a light quark or a gluon usually contains one.
We use the top-tagging dataset~\citep{kasieczka2019toptagging}, with top-quark jets as signal and jets from light quarks and gluons as background.
It contains 1.2 million training, 0.4 million validation and 0.4 million test jets.
To a tagger, each jet is a list of up to 200 particles, each given by its four-momentum, its energy and the three components of its momentum.
All students and the teachers trained from scratch take the same seven features per particle as input, and the two pretrained teachers keep the inputs of their pretraining (Appendix~\ref{app:inputs}).

\paragraph{Metrics}
A tagger gives each jet a score between 0 and 1, and jets with a score above a threshold are labelled as top jets.
The signal efficiency $\epsilon_S$ is the fraction of top jets that pass the threshold, and $\epsilon_B$ is the fraction of background jets that pass it.
The dataset is balanced, but in real collisions jets from light quarks and gluons far outnumber top jets, so a useful tagger must let through very few of them.
Following the benchmark~\citep{kasieczka2019toptagging}, we set the threshold so that $\epsilon_S=0.5$ or $0.3$ and report the background rejection $1/\epsilon_B$.
A rejection of 500 means that one background jet in 500 passes.
We also report the area under the receiver operating characteristic curve (AUC), the probability that a random top jet receives a higher score than a random background jet.
We evaluate every tagger on the full test set and refer to the performance of a teacher on these metrics as its strength.

\paragraph{Teachers}
The six teachers in Table~\ref{tab:teachers} span a range of strength, which lets us relate the benefit of distillation to the strength of the teacher.
Four are trained from scratch, and two are foundation models, pretrained and then fine-tuned on top tagging:
\begin{itemize}\setlength{\itemsep}{0pt}\setlength{\parskip}{0pt}
\item ParT~\citep{qu2022part}, a transformer that adds pair features to its attention scores
\item MIParT~\citep{wu2024mipart}, a transformer that also updates the pair features in additional attention layers
\item L-GATr~\citep{spinner2024lgatr}, a Lorentz-equivariant transformer
\item L-GATr-slim~\citep{petitjean2025economical}, a simpler and faster version of L-GATr
\item OmniLearned-S~\citep{bhimji2025omnilearned}, the small OmniLearned foundation model
\item ParT-f.t.~\citep{qu2022part}, ParT pretrained on JetClass and fine-tuned on top tagging
\end{itemize}

\begin{table}[t]
\centering
\caption{Teacher performance on the top-tagging test set: AUC and background rejection $1/\epsilon_B$ at two signal efficiencies $\epsilon_S$. We mark Lorentz-equivariant models with an asterisk and pretrained models with a dagger, and give means and standard deviations over three runs.}
\label{tab:teachers}
\vspace{4pt}
\small
\begin{tabular}{lrlrr}
\toprule
Teacher & Params & AUC & $1/\epsilon_B$ ($\epsilon_S{=}0.5$) & $1/\epsilon_B$ ($\epsilon_S{=}0.3$) \\
\midrule
ParT & 2.1M & 0.98571\std{0.00011} & 394\std{14} & 1{,}586\std{19} \\
MIParT & 0.7M & 0.98632\std{0.00005} & 467\std{5} & 1{,}871\std{69} \\
L-GATr\eqv{} & 1.1M & 0.98680\std{0.00005} & 528\std{14} & 2{,}189\std{176} \\
L-GATr-slim\eqv{} & 2.0M & 0.98711\std{0.00008} & 541\std{5} & 2{,}321\std{46} \\
\midrule
OmniLearned-S\pre{} & 2.7M & 0.98746\std{0.00001} & 546\std{11} & 2{,}403\std{100} \\
ParT-f.t.\pre{} & 2.1M & 0.98769\std{0.00005} & 644\std{16} & 2{,}556\std{0.2} \\
\bottomrule
\end{tabular}

\end{table}

\paragraph{Students}
The five students in Table~\ref{tab:students} have about 200 thousand parameters each, roughly a tenth of a typical teacher:
\begin{itemize}\setlength{\itemsep}{0pt}\setlength{\parskip}{0pt}
\item Transformer, a standard transformer over particles
\item LorentzNet~\citep{gong2022lorentznet}, a Lorentz-equivariant graph network that passes messages between all pairs of particles
\item ParT, a smaller version of the ParT teacher
\item MIParT, a smaller version of the MIParT teacher
\item L-GATr-slim, a smaller version of the L-GATr-slim teacher
\end{itemize}
Appendix~\ref{app:arch} gives the configuration of every model.

\begin{table}[t]
\centering
\caption{Students trained without a teacher: AUC and background rejection $1/\epsilon_B$ at $\epsilon_S=0.5$ and $0.3$, with means and standard deviations over three runs.}
\label{tab:students}
\vspace{4pt}
\small
\begin{tabular}{lrlrr}
\toprule
Student & Params & AUC & $1/\epsilon_B$ ($\epsilon_S{=}0.5$) & $1/\epsilon_B$ ($\epsilon_S{=}0.3$) \\
\midrule
Transformer & 188k & 0.98512\std{0.00013} & 393\std{5} & 1{,}612\std{121} \\
LorentzNet\eqv{} & 203k & 0.98556\std{0.00010} & 413\std{16} & 1{,}714\std{83} \\
ParT & 205k & 0.98565\std{0.00004} & 384\std{7} & 1{,}603\std{76} \\
MIParT & 200k & 0.98580\std{0.00019} & 410\std{5} & 1{,}767\std{83} \\
L-GATr-slim\eqv{} & 202k & 0.98622\std{0.00016} & 425\std{20} & 1{,}704\std{80} \\
\bottomrule
\end{tabular}

\end{table}

\paragraph{Distillation}
A tagger outputs a logit $z$ per jet, and $\sigma(z)$ is its predicted probability that the jet is signal.
A student trained without a teacher minimizes only the binary cross-entropy (BCE) to the hard label $y$, with the same recipe as the distilled student.
A distilled student also learns from the soft label, the prediction of the teacher softened by a temperature $T$, with the standard distillation loss~\citep{hinton2015distilling}
\begin{equation}
  \mathcal{L} = (1-\alpha)\,\mathrm{BCE}\big(\sigma(z_s),\,y\big) + \alpha\,T^{2}\,\mathrm{KL}\big(\sigma(z_t/T)\,\big\|\,\sigma(z_s/T)\big),
  \label{eq:kd}
\end{equation}
where $z_s$ and $z_t$ are the student and teacher logits, and the Kullback-Leibler (KL) divergence is taken between the Bernoulli distributions over signal and background that these probabilities define.
We fix $\alpha=0.7$ and $T=2$ in advance and use them for every pair of teacher and student, so that differences between results come only from the teacher and the student.
The teacher logits for the training jets are computed once and stored, so no teacher runs during student training.

\paragraph{Training}
We take the training recipe of each model from the literature and give every recipe with its source in Appendix~\ref{app:recipes}.
Students keep the checkpoint with the lowest binary cross-entropy to the labels on the validation set, with or without a teacher.
Each student of Section~\ref{sec:matrix} is trained three times, and Appendix~\ref{app:runs} describes how student and teacher runs are paired.

\section{Choice of teacher}
\label{sec:matrix}

We distil each of the six teachers into each of the five students.
Table~\ref{tab:matrix_auc} gives the AUC of every pair and Figure~\ref{fig:rej_by_teacher} its background rejection, with the values in Table~\ref{tab:matrix_rej} in Appendix~\ref{app:matrix_rej}.

\begin{figure}[t]
\centering
\includegraphics[width=\linewidth]{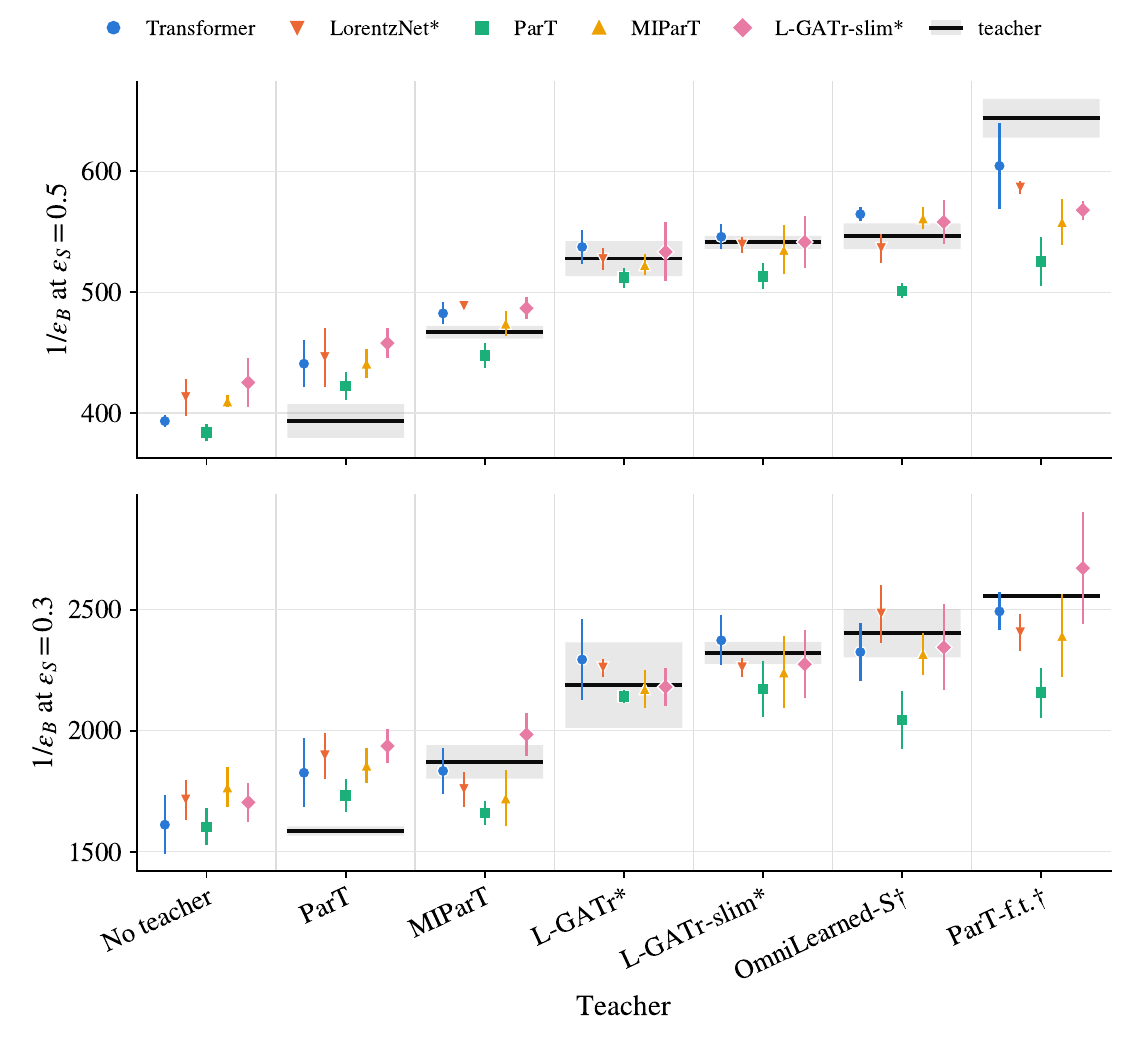}
\caption{Background rejection $1/\epsilon_B$ at $\epsilon_S=0.5$ (top) and $\epsilon_S=0.3$ (bottom) of every student, trained without a teacher and distilled from each teacher. Black lines and grey bands show the teachers. We show means and standard deviations over three runs.}
\label{fig:rej_by_teacher}
\end{figure}

\begin{table}[t]
\centering
\caption{AUC of every student (columns), trained without a teacher and distilled from each teacher (rows), with means and standard deviations over three runs.}
\label{tab:matrix_auc}
\vspace{4pt}
{\footnotesize
\setlength{\tabcolsep}{3.4pt}
\begin{tabular}{llllll}
\toprule
& \multicolumn{5}{c}{Student} \\ \cmidrule(lr){2-6}
Teacher & Transformer & LorentzNet\eqv{} & ParT & MIParT & L-GATr-slim\eqv{} \\
\midrule
No teacher & 0.98512\std{0.00013} & 0.98556\std{0.00010} & 0.98565\std{0.00004} & 0.98580\std{0.00019} & 0.98622\std{0.00016} \\
\midrule
ParT & 0.98617\std{0.00010} & 0.98630\std{0.00010} & 0.98607\std{0.00001} & 0.98624\std{0.00005} & 0.98643\std{0.00005} \\
MIParT & 0.98646\std{0.00007} & 0.98655\std{0.00010} & 0.98622\std{0.00008} & 0.98631\std{0.00002} & 0.98660\std{0.00001} \\
L-GATr\eqv{} & 0.98680\std{0.00011} & 0.98671\std{0.00018} & 0.98659\std{0.00007} & 0.98675\std{0.00007} & 0.98693\std{0.00001} \\
L-GATr-slim\eqv{} & 0.98691\std{0.00006} & 0.98685\std{0.00014} & 0.98671\std{0.00011} & 0.98690\std{0.00011} & 0.98712\std{0.00007} \\
\midrule
OmniLearned-S\pre{} & 0.98705\std{0.00007} & \textbf{0.98713}\std{0.00018} & 0.98677\std{0.00010} & 0.98705\std{0.00010} & 0.98727\std{0.00002} \\
ParT-f.t.\pre{} & \textbf{0.98722}\std{0.00005} & 0.98712\std{0.00009} & \textbf{0.98679}\std{0.00008} & \textbf{0.98711}\std{0.00007} & \textbf{0.98732}\std{0.00006} \\
\bottomrule
\end{tabular}
}

\end{table}

\begin{figure}[t]
\centering
\includegraphics[width=\linewidth]{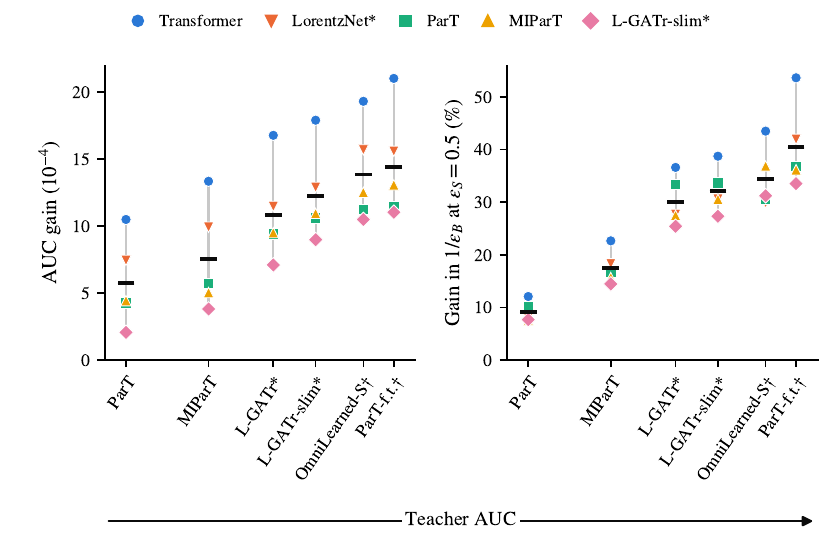}
\caption{Gain of every student from distillation against the AUC of its teacher, in AUC (left) and relative background rejection $1/\epsilon_B$ at $\epsilon_S=0.5$ (right). Black ticks show the mean over the five students. All values are means over three runs.}
\label{fig:strength}
\end{figure}

\paragraph{Gain for every student}
All 30 distilled students have a higher AUC and a higher background rejection at $\epsilon_S=0.5$ than the same student trained without a teacher.
At $\epsilon_S=0.3$, every distilled student improves except MIParT distilled from MIParT.
Weaker students gain more, since a strong student has less room to improve.
Transformer starts from the lowest AUC and gains the most from every teacher, in AUC and in background rejection at $\epsilon_S=0.5$.
Distilled from ParT-f.t., its background rejection at $\epsilon_S=0.5$ rises by more than half.
L-GATr-slim starts from the highest AUC and gains the least AUC from every teacher (Figure~\ref{fig:strength}).

\paragraph{Outperforming a larger model}
With a strong teacher, every student overtakes the L-GATr-slim student trained without a teacher, and so reaches the performance of a stronger architecture of the same size.
Most students distilled from the strongest teachers also outperform the L-GATr teacher, which has five times their parameters.
Transformer, LorentzNet and L-GATr-slim distilled from ParT-f.t.\ outperform even the L-GATr-slim teacher, which has ten times their parameters.

\paragraph{Effect of teacher strength}
Figure~\ref{fig:strength} shows the gain of every student against the AUC of its teacher, and stronger teachers give larger gains.
The mean gain in background rejection at $\epsilon_S=0.5$ grows from 9 percent with ParT to 40 percent with ParT-f.t.
For every student, a stronger teacher gives a higher AUC (Table~\ref{tab:matrix_auc}).
The only exception is LorentzNet, which reaches the same AUC with OmniLearned-S and with ParT-f.t.
Sharing an architecture or a symmetry with the teacher does not change this order.
Both Lorentz-equivariant students, for example, learn more from ParT-f.t., which has no built-in symmetry, than from either Lorentz-equivariant teacher.

\paragraph{Effect of pretraining}
Our two strongest teachers are both pretrained, and both fit the order of Figure~\ref{fig:strength}.
OmniLearned-S lies between L-GATr-slim and ParT-f.t.\ in strength, and so does the mean gain of its students at both signal efficiencies.
Because pretraining and strength go together in these teachers, we also distil from the ensemble of the three L-GATr-slim runs, a teacher trained from scratch that matches OmniLearned-S in AUC (Appendix~\ref{app:ensembles}).
Its students outperform those of a single L-GATr-slim and perform about as well as those of OmniLearned-S (Table~\ref{tab:ens_students}).
Pretraining therefore helps the student mainly by making the teacher stronger.

\paragraph{Weak teachers}
A weak teacher still improves its students, and some of them then outperform it.
Every student distilled from ParT exceeds it in AUC and in both background rejections, and three of the five students distilled from MIParT exceed its AUC.
ParT is weaker than the MIParT and L-GATr-slim students trained without a teacher (Tables~\ref{tab:teachers} and~\ref{tab:students}), and it still improves both.
A student can outperform its teacher because it also learns from the hard labels, while the soft labels regularize it much like label smoothing~\citep{yuan2020revisiting}.
The same happens when student and teacher have the same size~\citep{furlanello2018born}.

\section{Student size}
\label{sec:scaling}

\begin{figure}[t]
\centering
\includegraphics[width=\linewidth]{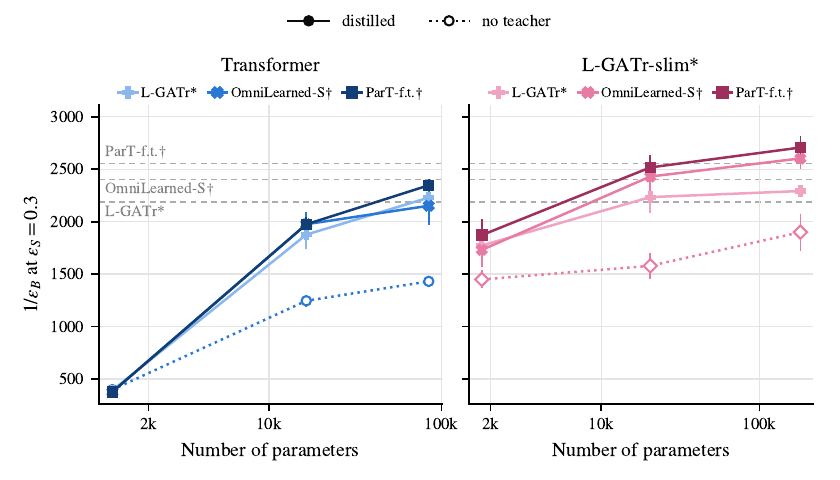}
\caption{Width study: background rejection $1/\epsilon_B$ at $\epsilon_S=0.3$ of ten-block students of decreasing width, trained without a teacher (dotted, hollow) and distilled from three teachers (solid, filled). Darker lines mark stronger teachers, and grey dashed lines show the teachers. Error bars show standard deviations over three runs.}
\label{fig:width}
\end{figure}

\begin{figure}[t]
\centering
\includegraphics[width=\linewidth]{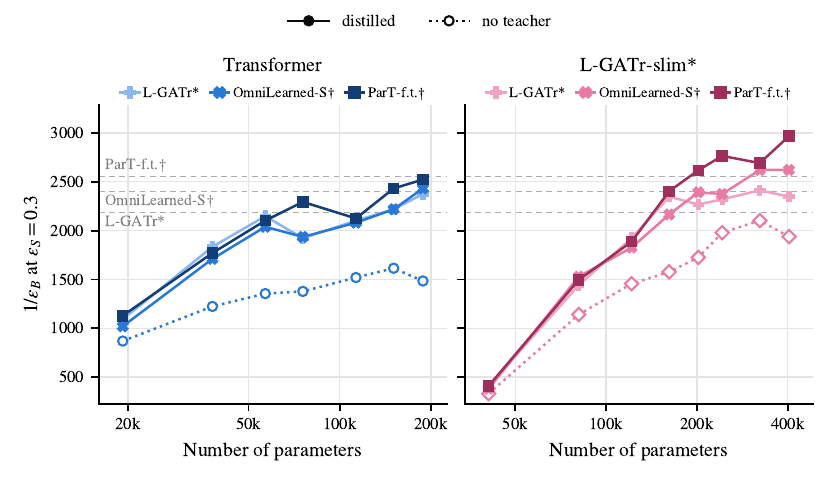}
\caption{Depth study: background rejection $1/\epsilon_B$ at $\epsilon_S=0.3$ of students of fixed width with a varying number of blocks, trained without a teacher (dotted, hollow) and distilled from three teachers (solid, filled), one run each. Darker lines mark stronger teachers, and grey dashed lines show the teachers.}
\label{fig:depth}
\end{figure}

Next, we ask how the benefit of distillation depends on the size of the student.
For this, we take our two strongest distilled students, which are also among the cheapest to run, Transformer and L-GATr-slim.
First, we scale them in width, using published ten-block networks at three widths, from about one thousand to 181 thousand parameters~\citep{petitjean2025economical}.
Then we scale them in depth, keeping the width of the students of Section~\ref{sec:matrix} and varying the number of blocks.
Every variant is trained without a teacher and distilled from L-GATr, OmniLearned-S and ParT-f.t., three strong teachers with very different architectures.
In the rest of the paper, we refer to these experiments as the width study and the depth study.
Figures~\ref{fig:width} and~\ref{fig:depth} show the background rejection at $\epsilon_S=0.3$, and Table~\ref{tab:size_short} and Appendix~\ref{app:results} give the full results.

\begin{table}[t]
\centering
\caption{Width study: AUC and background rejection $1/\epsilon_B$ at $\epsilon_S=0.5$ and $0.3$ of ten-block students of three widths, trained without a teacher and distilled from ParT-f.t., with means and standard deviations over three runs. Table~\ref{tab:size_other} gives the other two teachers.}
\label{tab:size_short}
\vspace{4pt}
{\footnotesize
\setlength{\tabcolsep}{3.5pt}
\begin{tabular}{lrlrrlrr}
\toprule
& & \multicolumn{3}{c}{No teacher} & \multicolumn{3}{c}{ParT-f.t.\pre{}} \\ \cmidrule(lr){3-5} \cmidrule(lr){6-8}
& & & \multicolumn{2}{c}{$1/\epsilon_B$} & & \multicolumn{2}{c}{$1/\epsilon_B$} \\ \cmidrule(lr){4-5} \cmidrule(lr){7-8}
Student & Params & AUC & $\epsilon_S{=}0.5$ & $\epsilon_S{=}0.3$ & AUC & $\epsilon_S{=}0.5$ & $\epsilon_S{=}0.3$ \\
\midrule
Transformer & 1.2k & 0.97422\std{0.00052} & 115\std{4} & 398\std{27} & 0.97401\std{0.00049} & 112\std{5} & 377\std{11} \\
 & 16k & 0.98388\std{0.00012} & 318\std{14} & 1{,}246\std{57} & 0.98624\std{0.00019} & 488\std{8} & 1{,}975\std{67} \\
 & 84k & 0.98495\std{0.00012} & 373\std{8} & 1{,}431\std{46} & 0.98688\std{0.00009} & 557\std{18} & 2{,}347\std{57} \\
\midrule
L-GATr-slim\eqv{} & 1.8k & 0.98491\std{0.00018} & 358\std{5} & 1{,}449\std{86} & 0.98567\std{0.00004} & 416\std{22} & 1{,}872\std{153} \\
 & 20k & 0.98539\std{0.00047} & 393\std{26} & 1{,}579\std{122} & 0.98729\std{0.00021} & 596\std{38} & 2{,}517\std{122} \\
 & 181k & 0.98635\std{0.00024} & 482\std{28} & 1{,}900\std{178} & 0.98772\std{0.00003} & 640\std{14} & 2{,}707\std{109} \\
\bottomrule
\end{tabular}
}

\end{table}

\paragraph{Effect of width}
A distilled student outperforms the next wider student of its architecture trained without a teacher, which has five to twelve times its parameters.
This holds in AUC and in both background rejections, down to 16 thousand parameters for Transformer and two thousand for L-GATr-slim.

\paragraph{Effect of depth}
Without a teacher, Transformer improves little beyond three blocks, so it stays far below the teachers and gains from each of them at every depth from one block.
Distilled from any of the three teachers, Transformer with two blocks outperforms Transformer with ten blocks trained without a teacher.
L-GATr-slim keeps improving with depth on its own and reaches the AUC of the L-GATr teacher with ten blocks, so its gain from L-GATr shrinks.
OmniLearned-S and ParT-f.t.\ stay stronger than L-GATr-slim at every depth, so they keep improving it.

\paragraph{Beyond the state of the art}
Distilled from ParT-f.t., L-GATr-slim with 20 thousand parameters outperforms every tagger trained from scratch, both our teachers and the published taggers~\citep{petitjean2025economical}, which have up to 150 times its parameters.
L-GATr and OmniLearned-S help at this size as well, and raise the background rejection of both students with about 20 thousand parameters by about half.

\paragraph{Architecture of small students}
Without a teacher, L-GATr-slim with two thousand parameters performs about as well as Transformer with 84 thousand (Table~\ref{tab:size_short}), in line with earlier results~\citep{petitjean2025economical}.
At 200 thousand parameters, distillation from a strong teacher closes the gap between the two architectures (Section~\ref{sec:matrix}), and at the smallest size only L-GATr-slim gains from it.
The choice of architecture therefore matters most for small students.

\paragraph{Limits of distillation}
At the smallest width, distillation reaches its limit for Transformer.
With one thousand parameters, Transformer does not improve with any teacher, because its capacity limits what it can learn (Section~\ref{sec:learn}).
L-GATr-slim with two thousand parameters still improves with every teacher.
Distilled from ParT-f.t., it matches the AUC of ParT, which has more than a thousand times its parameters, and exceeds its background rejections.

\section{What the student learns from its teacher}
\label{sec:learn}

\begin{figure}[t]
\centering
\includegraphics[width=\linewidth]{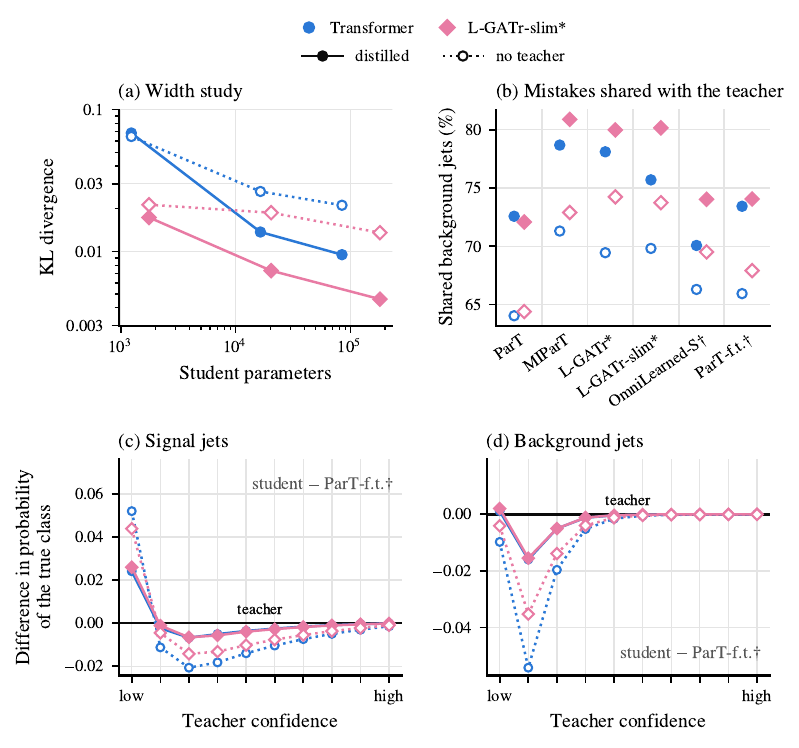}
\caption{What Transformer and L-GATr-slim learn from their teacher, distilled (solid, filled) and trained without a teacher (dotted, hollow). (a)~KL divergence from ParT-f.t.\ to the student against the number of student parameters in the width study. (b)~Percentage of the background jets each teacher accepts at $\epsilon_S=0.5$ that the student also accepts. (c, d)~Probability the student assigns to the true class minus that of ParT-f.t., for signal (c) and background (d) jets, in ten groups of equal size ordered by the probability ParT-f.t.\ assigns to the true class. All values are means over three runs.}
\label{fig:learn}
\end{figure}

Sections~\ref{sec:matrix} and~\ref{sec:scaling} measure how much a student gains from its teacher.
Here we ask what the student learns from the teacher.
Every tagger assigns each jet a probability of being a top jet.
We compare these predicted probabilities on the test jets for the teacher, the distilled student and the same student trained without a teacher.
We measure how closely the student reproduces the probabilities of the teacher, which mistakes it shares with the teacher and how its agreement with the teacher depends on how confident the teacher is.
Figure~\ref{fig:learn} shows these analyses for Transformer and L-GATr-slim, and Appendix~\ref{app:learn} gives every student.

\paragraph{Agreement with the teacher}
We measure how closely a student reproduces its teacher, its fidelity~\citep{stanton2021does}, by the KL divergence from the predicted probabilities of the teacher to those of the student.
We average it over the test jets.
It is zero when the student assigns every jet the same probability as the teacher.
Distillation lowers this divergence for all 30 pairs of teacher and student of Section~\ref{sec:matrix}, by more than half on average.
Figure~\ref{fig:learn}a shows the width study with ParT-f.t.\ as teacher, where distillation lowers it down to 16 thousand parameters.
Below this size, the decrease is much smaller.
Transformer with one thousand parameters is as far from its teacher with distillation as without it.
L-GATr-slim with two thousand parameters still moves a little closer to its teacher, which matches its gain from distillation at this size, where Transformer gains nothing (Section~\ref{sec:scaling}).
A large capacity gap between teacher and student is known to limit distillation~\citep{cho2019efficacy,mirzadeh2020ta}.
The other two teachers of the width study show the same pattern, and in the depth study the shallowest students move least towards their teacher (Figure~\ref{fig:learn_size}).

\paragraph{Mistakes inherited from the teacher}
A student can also inherit the mistakes of its teacher~\citep{ojha2023knowledge}.
For example, a jet tagger can incorrectly accept a background jet as a top jet, a false positive.
We count these mistakes with every tagger working at a signal efficiency of $\epsilon_S=0.5$.
A distilled student repeats more of the mistakes of its teacher than the same student trained without a teacher, for all 30 pairs (Figures~\ref{fig:learn}b and~\ref{fig:learn_shared}).
When two teachers disagree on a background jet, the student is more likely to accept it if it was distilled from the teacher that accepts it (Figure~\ref{fig:learn_distinctive}).
The mistakes a student inherits are therefore those of its own teacher.

\paragraph{Mistakes removed by distillation}
Inheriting these mistakes costs the student little, because distillation removes many more mistakes than it adds.
On the background jets the teacher rejects, a distilled student makes far fewer mistakes than without a teacher.
On those the teacher accepts it makes somewhat more, but never as many as it removes (Figure~\ref{fig:learn_accepted} and Table~\ref{tab:accepted}).
The student thus learns from the jets its teacher classifies correctly, and this is where its gain in background rejection comes from.

\paragraph{Agreement by teacher confidence}
Finally, we ask on which jets the student follows its teacher.
We define the confidence of the teacher on a jet as the probability it assigns to the true class, and split the jets of each class into ten groups of equal size, from the least to the most confident.
In the most confident groups, every student already agrees with the teacher.
In the others, distillation moves the student towards the teacher (Figure~\ref{fig:learn}c and d, and Figures~\ref{fig:learn_perjet}, \ref{fig:learn_perjet_omni} and~\ref{fig:learn_perjet_lgatr} for every student and teacher).
This helps the student wherever the teacher is closer to the true class, which is almost everywhere.
The exception is the least confident group of top jets, about half of which the teacher takes for background jets.
There the student trained without a teacher does better than the teacher, and distillation pulls it towards the mistakes of the teacher.
After distillation, Transformer and L-GATr-slim make almost the same predictions in every group, because both follow the same teacher.

\section{Inference cost}
\label{sec:inference}

We now measure the inference cost of every teacher and student and compare them at fixed cost~\citep{favaro2026virtues}.

\paragraph{Protocol}
We measure inference cost on one NVIDIA RTX 6000 Ada GPU and one Intel Xeon Platinum 8352Y CPU, with PyTorch 2.11 and CUDA 12.8.
On the GPU we report the throughput at a batch of 512 jets and the peak memory, and on the CPU the latency for a single jet on one core.
Every model runs in full 32-bit precision, as in training, and we report its fastest execution setting (Appendix~\ref{app:benchmark}).
A timed call starts from the four-momenta of the jets in CPU memory and includes the input features and the network.
Each measurement runs in a fresh process and takes the median of 100 calls after at least 100 warm-up calls.
We repeat it three times and report the median.
Our protocol extends an earlier cost benchmark of equivariant transformers~\citep{favaro2026virtues}, and Appendix~\ref{app:benchmark} repeats all measurements on jets with exactly 50 particles, as in that benchmark.

\begin{table}[t]
\centering
\caption{Performance and inference cost on real test jets in the fastest 32-bit setting: operations per jet in millions (median over 300 jets, one multiply-accumulate counted as two), GPU throughput and peak memory at a batch of 512 jets and CPU latency for one jet on one core. Teachers are shown with their own performance and students after distillation from ParT-f.t., with standard deviations of the background rejection over three runs. Deep Sets + GNN~\citep{benyas2026accelerators} is distilled from OmniLearned-L, and its performance is as published. Appendix~\ref{app:cost_full} gives the same table without a teacher and for every other teacher.}
\label{tab:cost}
\vspace{4pt}
{\small\setlength{\tabcolsep}{4pt}
\begin{tabular}{lrrrrrrrr}
\toprule
 & & & \multicolumn{2}{c}{$1/\epsilon_B$} & & \multicolumn{2}{c}{GPU} & CPU \\ \cmidrule(lr){4-5} \cmidrule(lr){7-8}
Model & Params & AUC & $\epsilon_S{=}0.5$ & $\epsilon_S{=}0.3$ & MFLOP & jets/s & MB & ms/jet \\
\midrule
\multicolumn{9}{l}{\textit{Teachers}} \\
ParT & 2.1M & 0.98571 & 394\std{14} & 1{,}586\std{19} & 197 & 18{,}198 & 1{,}099 & 7.9 \\
MIParT & 0.7M & 0.98632 & 467\std{5} & 1{,}871\std{69} & 76.0 & 7{,}225 & 5{,}132 & 8.9 \\
L-GATr\eqv{} & 1.1M & 0.98680 & 528\std{14} & 2{,}189\std{176} & 1{,}868 & 4{,}263 & 707 & 50.5 \\
L-GATr-slim\eqv{} & 2.0M & 0.98711 & 541\std{5} & 2{,}321\std{46} & 309 & 16{,}546 & 339 & 8.1 \\
OmniLearned-S\pre{} & 2.7M & 0.98746 & 546\std{11} & 2{,}403\std{100} & 427 & 3{,}736 & 13{,}221 & 17.0 \\
ParT-f.t.\pre{} & 2.1M & 0.98769 & 644\std{16} & 2{,}556\std{0.2} & 197 & 18{,}286 & 1{,}099 & 8.0 \\
\midrule
\multicolumn{9}{l}{\textit{Students, distilled from ParT-f.t.\pre{}}} \\
Transformer & 188k & 0.98722 & 604\std{35} & 2{,}493\std{78} & 21.6 & 70{,}175 & 87 & 2.7 \\
LorentzNet\eqv{} & 203k & 0.98712 & 586\std{5} & 2{,}405\std{77} & 537 & 7{,}601 & 1{,}722 & 11.0 \\
ParT & 205k & 0.98679 & 525\std{21} & 2{,}157\std{104} & 21.9 & 26{,}228 & 2{,}166 & 2.3 \\
MIParT & 200k & 0.98711 & 558\std{19} & 2{,}393\std{172} & 20.4 & 12{,}181 & 1{,}709 & 7.8 \\
L-GATr-slim\eqv{} & 202k & 0.98732 & 568\std{8} & 2{,}672\std{233} & 29.7 & 65{,}308 & 150 & 2.6 \\
\midrule
\multicolumn{9}{l}{\textit{Width study, distilled from ParT-f.t.\pre{}}} \\
Transformer & 84k & 0.98688 & 557\std{18} & 2{,}347\std{57} & 10.6 & 69{,}945 & 70 & 2.3 \\
Transformer & 16k & 0.98624 & 488\std{8} & 1{,}975\std{67} & 2.87 & 115{,}996 & 112 & 2.0 \\
Transformer & 1.2k & 0.97401 & 112\std{5} & 377\std{11} & 0.45 & 112{,}829 & 149 & 1.5 \\
L-GATr-slim\eqv{} & 181k & 0.98772 & 640\std{14} & 2{,}707\std{109} & 28.6 & 40{,}269 & 115 & 3.6 \\
L-GATr-slim\eqv{} & 20k & 0.98729 & 596\std{38} & 2{,}517\std{122} & 5.70 & 76{,}488 & 161 & 2.8 \\
L-GATr-slim\eqv{} & 1.8k & 0.98567 & 416\std{22} & 1{,}872\std{153} & 1.50 & 77{,}262 & 111 & 2.2 \\
\midrule
\multicolumn{9}{l}{\textit{Distilled from OmniLearned-L\pre{}~\citep{benyas2026accelerators}}} \\
Deep Sets + GNN & 17k & 0.9859 & 450 & 1{,}980 & 28.6 & 63{,}877 & 1{,}074 & 1.4 \\
\bottomrule
\end{tabular}
}

\end{table}

\begin{figure}[t]
\centering
\includegraphics[width=\linewidth]{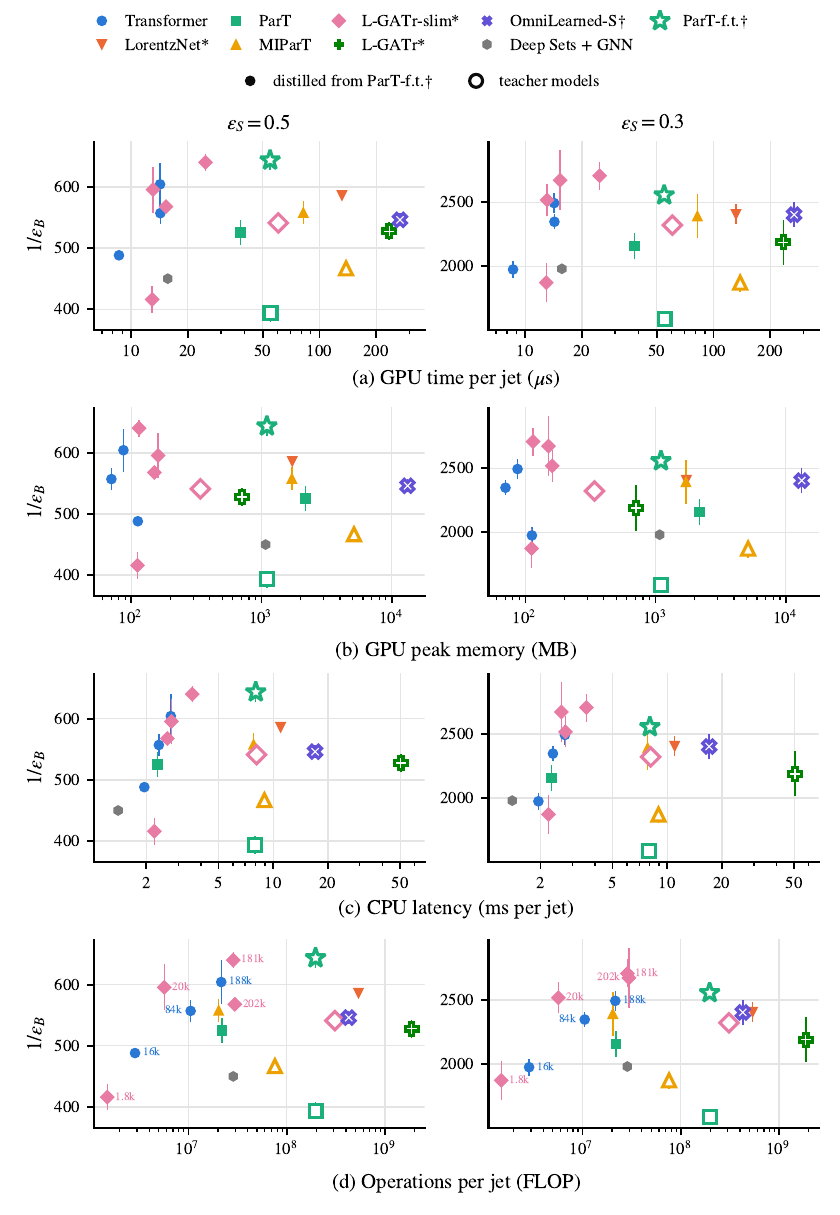}
\caption{Background rejection $1/\epsilon_B$ at $\epsilon_S=0.5$ (left) and $\epsilon_S=0.3$ (right) against (a)~GPU time per jet and (b)~GPU peak memory at a batch of 512 jets, (c)~CPU latency for one jet and (d)~operations per jet, in the fastest 32-bit setting. Students distilled from ParT-f.t.\ are filled and teachers hollow, and (d) gives the sizes of the Transformer and L-GATr-slim students. Error bars show standard deviations over three runs.}
\label{fig:cost_perf}
\end{figure}

\paragraph{Operations per jet}
Table~\ref{tab:cost} gives the operations per jet of every model, counted with PyTorch's \texttt{FlopCounterMode} as in~\citep{spinner2025lloca}.
This count includes matrix multiplications, convolutions and attention but not elementwise operations.
Distilled L-GATr-slim with 20 thousand parameters and Transformer with 188 thousand both outperform every teacher trained from scratch with a fraction of their operations.
Our strongest student, L-GATr-slim with 181 thousand parameters, performs as well as our strongest teacher, ParT-f.t., with a seventh of its operations.
Transformer of about the same size needs even fewer operations and is only slightly weaker.

\paragraph{Cost of pair features}
The five students in Table~\ref{tab:cost} have about 200 thousand parameters each, yet their GPU throughput differs by almost an order of magnitude (Figure~\ref{fig:cost_params}).
ParT and MIParT need about as many operations per jet as Transformer but run several times slower, because they compute and store features for every pair of particles.
This cost grows with the square of the number of particles, while the parameter count stays fixed.
Attention also compares every pair, but fused attention kernels never store the full matrix~\citep{favaro2026virtues}.
LorentzNet and Deep Sets + GNN also work on all pairs of particles, and need far more memory than Transformer.

\paragraph{Performance at fixed cost}
Figure~\ref{fig:cost_perf} shows the background rejection of every model against its cost.
The models with the highest rejection at each cost form the Pareto front.
Among the students of the width study, Transformer forms this front at the shortest times on both devices and L-GATr-slim at longer times, as in earlier comparisons at fixed cost~\citep{favaro2026virtues}.
Distillation keeps this ordering and raises the front by about half in background rejection.
Deep Sets + GNN joins the front only for CPU latency, and its performance is below that of most of our students.
The strongest student matches the AUC of ParT-f.t.\ at twice its throughput.
Distilled from OmniLearned-S, L-GATr-slim with 20 thousand parameters rejects more background than its teacher and processes about twenty times as many jets per second (Table~\ref{tab:cost_omnilearned_s}).

\paragraph{Latency on the CPU}
On the CPU, which processes one jet at a time, the latency of the students hardly depends on their size (Figure~\ref{fig:cost_perf}c), so distillation gains less speed than on the GPU.
Deep Sets + GNN is the fastest model on the CPU, followed by Transformer with one and 16 thousand parameters.

\FloatBarrier
\section{Discussion}
\label{sec:discussion}

We have studied knowledge distillation for jet tagging with six teachers and five students, and in two scaling studies of the student size.
Distillation improved every student down to two thousand parameters, at no cost at inference.
The gain followed the strength of the teacher, and a teacher trained from scratch taught about as well as a pretrained teacher of similar strength.

In the width study, distilled L-GATr-slim with 20 thousand parameters outperformed every tagger trained from scratch at a fraction of the inference cost of every teacher.
At the smallest width, Transformer with one thousand parameters could not follow its teacher and gained nothing from distillation.
L-GATr-slim with two thousand parameters still gained, matched the AUC of the ParT teacher with a thousand times its parameters and exceeded its background rejections.
A small tagger should therefore combine a strong teacher with an architecture that stays strong at small size, such as L-GATr-slim.

To understand where the gain comes from, we compared the predictions of students and teachers.
The students followed their teachers closely, down to copying their mistakes, but distillation removed many more mistakes than it added.

Inference cost limits the use of taggers in experiments and is hard to predict from simple measures such as the parameter count, so we measured it directly on GPUs and CPUs.
Distilled Transformer and L-GATr-slim students formed almost all of the Pareto front of performance against inference cost, and distilled L-GATr-slim outperformed the state-of-the-art taggers trained from scratch at a fraction of their cost.

Our study uses one binary benchmark and one distillation loss, with the same settings for every pair of teacher and student.
Natural next steps are to tune these students for deployment, where optimized settings may improve them further, and to test them on larger datasets such as JetClass.
Larger foundation models, such as OmniLearned-L, could also be distilled into students such as L-GATr-slim.
Interpretability studies of jet taggers~\citep{agarwal2026lorentz,rai2026dissecting} could show what the students learn from their teachers beyond their predictions.

\section*{Acknowledgements}
We thank Jonas Spinner for helpful discussions and suggestions in the early stages of this work.
We thank Ahmed Hammad and Mihoko Nojiri for feedback on the results and the manuscript.
JA was supported by the MEXT Grant-in-Aid for Transformative Research Areas ``Foundation of Machine Learning Physics'' (Grant Number 22H05113) and by JST ARiSE (Grant Number JPMJAR2633), Japan.
We thank the Department of Computer Science and Information Systems, BITS Pilani, for access to its GPU server.

\section*{AI use statement}
In this work, we used AI tools for writing and debugging code, running experiments and minor grammar and sentence edits for the text.
We have reviewed all AI-assisted work.
We take responsibility for the final content of this work.

\clearpage
\bibliographystyle{plainnat}
\bibliography{refs}

\clearpage
\appendix

\FloatBarrier
\section{Models}
\label{app:models}

\subsection{Inputs}
\label{app:inputs}

Every student and every teacher trained from scratch takes seven features of each particle as input, computed relative to the jet axis.
These are the differences $\Delta\eta$ and $\Delta\phi$ in pseudorapidity and azimuth, the logarithms $\log p_T$ and $\log E$ of the transverse momentum and energy, the logarithms $\log(p_T/p_{T,\text{jet}})$ and $\log(E/E_\text{jet})$ of their fractions of the jet's values and the angular distance $\Delta R=\sqrt{\Delta\eta^2+\Delta\phi^2}$.
Models with pair interactions also compute four features for every pair of particles $a$ and $b$, as defined for ParT~\citep{qu2022part},
\begin{align}
  \ln\Delta, &\quad \Delta=\sqrt{(y_a-y_b)^2+(\phi_a-\phi_b)^2},\\
  \ln k_T, &\quad k_T=\min(p_{T,a},p_{T,b})\,\Delta,\\
  \ln z, &\quad z=\min(p_{T,a},p_{T,b})/(p_{T,a}+p_{T,b}),\\
  \ln m^2, &\quad m^2=(E_a+E_b)^2-\lVert\mathbf{p}_a+\mathbf{p}_b\rVert^2,
\end{align}
where $y$ is the rapidity.
The two pretrained teachers keep the inputs of their pretraining.
ParT-f.t.\ uses the same seven features and pair features, but only the 128 particles with the highest transverse momentum.
OmniLearned-S uses four of the seven features, $\Delta\eta$, $\Delta\phi$, $\log p_T$ and $\log E$, for up to 150 particles.
Before computing the features, we raise the energy of every particle with a mass below 5\,MeV so that its mass equals 5\,MeV, as in~\citep{spinner2025lloca}.
The Lorentz-equivariant models receive reference directions, which break the Lorentz symmetry down to the symmetry of the detector.
LorentzNet receives the two beam directions $(1,0,0,\pm1)$ and the time direction $(1,0,0,0)$, as in~\citep{spinner2025lloca}.
L-GATr and L-GATr-slim receive the time direction and three lightlike vectors along the beam axis and the two transverse axes, as in~\citep{favaro2026virtues}.

\subsection{Architectures}
\label{app:arch}

All students are smaller versions of published architectures (Table~\ref{tab:arch}).
The depth variants of Section~\ref{sec:scaling} keep the configuration of their student and change only the number of blocks.
ParT has eight particle-attention blocks followed by two class-attention blocks, and ParT-f.t.\ is the ParT network trained on JetClass~\citep{qu2022part}, with the same blocks. MIParT starts with blocks that also attend over pair features, and OmniLearned-S has eight transformer blocks and two blocks in its classification head.
Each model reads out the jet with a mean over particles, with learned class tokens or with the global token of L-GATr.

The ParT student keeps the inputs of ParT, the seven per-particle features and the four pair features that bias the attention.
It has half the width, half the heads and four of the ten blocks of ParT, reads out with a mean over particles and passes its pair features through a small network with one hidden layer of 32 channels.
This leaves about a tenth of the parameters of ParT.
Trained with the recipe of ParT, it reaches the AUC of ParT within the spread between runs (Tables~\ref{tab:teachers} and~\ref{tab:students}), in line with the finding that ParT keeps its performance with fewer parameters~\citep{cheng2026redundancy}.

Transformer widens the largest network of the width study, a published ten-block Transformer with 32 channels~\citep{petitjean2025economical}, to 48 channels, with six heads, an MLP of twice the width and no dropout.
In the depth study, Transformer with zero blocks embeds each particle with one linear layer and averages over particles, so it sees the jet only through this average.

\begin{table}[h]
\centering
\caption{Configuration of every model. Width counts hidden channels, as multivector (mv), vector (v) and scalar (s) channels for the L-GATr family; blocks counts the blocks of each model, with a sum for blocks of different kinds.}
\label{tab:arch}
\vspace{4pt}
{\small
\begin{tabular}{llllll}
\toprule
Model & Width & Heads & Blocks & Read-out & Params \\
\midrule
\multicolumn{6}{l}{\textit{Teachers}} \\
ParT & 128 & 8 & 8 + 2 & class token & 2{,}141{,}005 \\
MIParT & 64 & 8 & 5 + 5 + 2 & class token & 656{,}949 \\
L-GATr\eqv{} & 16 mv, 32 s & 8 & 12 & global token & 1{,}077{,}474 \\
L-GATr-slim\eqv{} & 32 v, 96 s & 8 & 12 & global token & 2{,}030{,}817 \\
OmniLearned-S\pre{} & 128 & 8 & 8 + 2 & 4 class tokens & 2{,}708{,}149 \\
ParT-f.t.\pre{} & 128 & 8 & 8 + 2 & class token & 2{,}141{,}562 \\
\midrule
\multicolumn{6}{l}{\textit{Students}} \\
Transformer & 48 & 6 & 10 & mean & 188{,}113 \\
LorentzNet\eqv{} & 68 & -- & 6 & mean & 202{,}511 \\
ParT & 64 & 4 & 4 & mean & 205{,}093 \\
MIParT & 32 & 8 & 6 + 8 + 2 & class token & 199{,}837 \\
L-GATr-slim\eqv{} & 12 v, 48 s & 4 & 5 & global token & 202{,}213 \\
\midrule
\multicolumn{6}{l}{\textit{Width study}} \\
Transformer & 32 & 4 & 10 & mean & 84{,}449 \\
Transformer & 16 & 2 & 10 & mean & 16{,}465 \\
Transformer & 4 & 1 & 10 & mean & 1{,}237 \\
L-GATr-slim\eqv{} & 8 v, 32 s & 4 & 10 & global token & 180{,}969 \\
L-GATr-slim\eqv{} & 4 v, 16 s & 2 & 10 & global token & 20{,}405 \\
L-GATr-slim\eqv{} & 2 v, 4 s & 1 & 10 & global token & 1{,}763 \\
\bottomrule
\end{tabular}
}

\end{table}

\FloatBarrier
\section{Training}
\label{app:training}

\subsection{Recipes}
\label{app:recipes}

Every model is trained with the published recipe of its architecture, and Table~\ref{tab:recipes} lists those of the students.
ParT and MIParT, as teachers and as students, use their published recipes~\citep{qu2022part,wu2024mipart} as implemented in LLoCa~\citep{spinner2025lloca}, and ParT-f.t.\ is fine-tuned with the recipe of ParT.
L-GATr uses its most recent published recipe~\citep{favaro2026virtues}, and L-GATr-slim, LorentzNet and OmniLearned-S use their original recipes~\citep{petitjean2025economical,gong2022lorentznet,bhimji2025omnilearned}.
The width study uses the published recipe for each size~\citep{petitjean2025economical}, Transformer the recipe of its largest network and the depth variants the recipe of their architecture.
All students clip gradients at a norm of 1 and train in 32-bit precision.
Table~\ref{tab:published} compares the teachers and LorentzNet trained with these recipes with the published performance of the same models.

\begin{table}[h]
\centering
\caption{Training recipes. Steps are optimizer updates.}
\label{tab:recipes}
\vspace{4pt}
{\small
\begin{tabular}{lllllrl}
\toprule
Model & Optimizer & Learning rate & Weight decay & Batch & Steps & Schedule \\
\midrule
Transformer & AdamW & $10^{-3}$ & 0.1 & 256 & 200{,}000 & cosine \\
LorentzNet\eqv{} & AdamW & $10^{-3}$ & $10^{-2}$ & 128 & 336{,}000 & cosine \\
ParT & Ranger & $10^{-3}$ & $10^{-2}$ & 512 & 48{,}000 & flat, then decay \\
MIParT & Ranger & $10^{-3}$ & $10^{-2}$ & 256 & 72{,}000 & flat, then decay \\
L-GATr-slim\eqv{} & Lion & $3\times10^{-5}$ & 2 & 128 & 200{,}000 & cosine \\
\midrule
Transformer 84k & AdamW & $10^{-3}$ & 0.1 & 256 & 200{,}000 & cosine \\
Transformer 16k & AdamW & $3\times10^{-3}$ & 0 & 512 & 200{,}000 & cosine \\
Transformer 1.2k & AdamW & $10^{-2}$ & 0 & 4{,}096 & 200{,}000 & cosine \\
L-GATr-slim\eqv{} 181k & AdamW & $10^{-3}$ & 0.1 & 256 & 200{,}000 & cosine \\
L-GATr-slim\eqv{} 20k & AdamW & $3\times10^{-3}$ & 0 & 512 & 200{,}000 & cosine \\
L-GATr-slim\eqv{} 1.8k & AdamW & $10^{-2}$ & 0 & 4{,}096 & 200{,}000 & cosine \\
\bottomrule
\end{tabular}
}

\end{table}

\begin{table}[h]
\centering
\caption{Performance of the teachers and of LorentzNet trained without a teacher, with means and standard deviations over three runs, and the published performance of the same models~\citep{qu2022part,wu2024mipart,spinner2024lgatr,petitjean2025economical,bhimji2025omnilearned,gong2022lorentznet}.}
\label{tab:published}
\vspace{4pt}
{\footnotesize
\setlength{\tabcolsep}{2.8pt}
\begin{tabular}{lrlrrrlrr}
\toprule
& \multicolumn{4}{c}{This work} & \multicolumn{4}{c}{Published} \\
\cmidrule(lr){2-5} \cmidrule(lr){6-9}
& & & \multicolumn{2}{c}{$1/\epsilon_B$} & & & \multicolumn{2}{c}{$1/\epsilon_B$} \\
\cmidrule(lr){4-5} \cmidrule(lr){8-9}
Model & Params & AUC & $\epsilon_S{=}0.5$ & $\epsilon_S{=}0.3$ & Params & AUC & $\epsilon_S{=}0.5$ & $\epsilon_S{=}0.3$ \\
\midrule
ParT & 2.1M & 0.98571\std{0.00011} & 394\std{14} & 1{,}586\std{19} & 2.1M & 0.9858 & 413\std{16} & 1{,}602\std{81} \\
MIParT & 0.7M & 0.98632\std{0.00005} & 467\std{5} & 1{,}871\std{69} & 0.7M & 0.9868 & 505\std{8} & 2{,}010\std{97} \\
L-GATr\eqv{} & 1.1M & 0.98680\std{0.00005} & 528\std{14} & 2{,}189\std{176} & 1.1M & 0.9870\std{0.0001} & 540\std{20} & 2{,}240\std{70} \\
L-GATr-slim\eqv{} & 2.0M & 0.98711\std{0.00008} & 541\std{5} & 2{,}321\std{46} & 2.0M & 0.9869\std{0.0001} & 546\std{7} & 2{,}264\std{93} \\
\midrule
OmniLearned-S\pre{} & 2.7M & 0.98746\std{0.00001} & 546\std{11} & 2{,}403\std{100} & 3M & 0.9875 & 565\std{12} & 2{,}637\std{128} \\
ParT-f.t.\pre{} & 2.1M & 0.98769\std{0.00005} & 644\std{16} & 2{,}556\std{0.2} & 2.1M & 0.9877 & 691\std{15} & 2{,}766\std{130} \\
\midrule
LorentzNet\eqv{} & 203k & 0.98556\std{0.00010} & 413\std{16} & 1{,}714\std{83} & 224k & 0.9868 & 498\std{18} & 2{,}195\std{173} \\
\bottomrule
\end{tabular}
}
\end{table}

\subsection{Runs and seeds}
\label{app:runs}

Every student of Section~\ref{sec:matrix} and of the width study is trained with three seeds, and the depth study uses one run per cell.
ParT, L-GATr and L-GATr-slim are also trained three times as teachers, and student run $k$ is distilled from teacher run $k$.
MIParT, ParT-f.t.\ and OmniLearned-S are trained three times as well, and Table~\ref{tab:teachers} reports their mean, but their students are all distilled from their first run.
Training the 248 student runs took about 1,500 hours on NVIDIA RTX 6000 Ada GPUs.

\subsection{Teacher ensembles}
\label{app:ensembles}

Averaging the logits of the three runs of a teacher raises its AUC for every teacher (Table~\ref{tab:ensembles}).
The gain is larger for the teachers trained from scratch, whose runs differ more, than for the two pretrained teachers, whose runs start from the same pretrained weights.
The ensemble of the three L-GATr-slim runs matches OmniLearned-S in AUC and rejects more background, and Table~\ref{tab:ens_students} compares their students.

\begin{table}[h]
\centering
\caption{AUC and background rejection $1/\epsilon_B$ at $\epsilon_S=0.5$ and $0.3$ of every teacher as a single run, with the mean and standard deviation over three runs, and as the ensemble of its three runs, which averages their logits.}
\label{tab:ensembles}
\vspace{4pt}
{\small\setlength{\tabcolsep}{4pt}
\begin{tabular}{llrrlrr}
\toprule
& \multicolumn{3}{c}{Single run} & \multicolumn{3}{c}{Ensemble of three runs} \\
\cmidrule(lr){2-4} \cmidrule(lr){5-7}
& & \multicolumn{2}{c}{$1/\epsilon_B$} & & \multicolumn{2}{c}{$1/\epsilon_B$} \\
\cmidrule(lr){3-4} \cmidrule(lr){6-7}
Teacher & AUC & $\epsilon_S{=}0.5$ & $\epsilon_S{=}0.3$ & AUC & $\epsilon_S{=}0.5$ & $\epsilon_S{=}0.3$ \\
\midrule
ParT & 0.98571\std{0.00011} & 394\std{14} & 1{,}586\std{19} & 0.98630 & 448 & 1{,}960 \\
MIParT & 0.98632\std{0.00005} & 467\std{5} & 1{,}871\std{69} & 0.98663 & 499 & 1{,}972 \\
L-GATr\eqv{} & 0.98680\std{0.00005} & 528\std{14} & 2{,}189\std{176} & 0.98719 & 554 & 2{,}462 \\
L-GATr-slim\eqv{} & 0.98711\std{0.00008} & 541\std{5} & 2{,}321\std{46} & 0.98748 & 589 & 2{,}490 \\
\midrule
OmniLearned-S\pre{} & 0.98746\std{0.00001} & 546\std{11} & 2{,}403\std{100} & 0.98755 & 541 & 2{,}645 \\
ParT-f.t.\pre{} & 0.98769\std{0.00005} & 644\std{16} & 2{,}556\std{0.2} & 0.98776 & 665 & 2{,}492 \\
\bottomrule
\end{tabular}
}
\end{table}

\begin{table}[h]
\centering
\caption{AUC and background rejection $1/\epsilon_B$ at $\epsilon_S=0.5$ and $0.3$ of every student of Section~\ref{sec:matrix} distilled from L-GATr-slim, from the ensemble of its three runs (L-GATr-slim ens.) and from OmniLearned-S, with means and standard deviations over three runs.}
\label{tab:ens_students}
\vspace{4pt}
{\footnotesize
\setlength{\tabcolsep}{3pt}
\begin{tabular}{llllll}
\toprule
& \multicolumn{5}{c}{Student} \\ \cmidrule(lr){2-6}
Teacher & Transformer & LorentzNet\eqv{} & ParT & MIParT & L-GATr-slim\eqv{} \\
\midrule
\multicolumn{6}{l}{\textit{AUC}} \\
L-GATr-slim\eqv{} & 0.98691\std{0.00006} & 0.98685\std{0.00014} & 0.98671\std{0.00011} & 0.98690\std{0.00011} & 0.98712\std{0.00007} \\
L-GATr-slim\eqv{} ens. & 0.98701\std{0.00003} & 0.98711\std{0.00009} & 0.98678\std{0.00010} & 0.98698\std{0.00010} & 0.98719\std{0.00003} \\
OmniLearned-S\pre{} & 0.98705\std{0.00007} & 0.98713\std{0.00018} & 0.98677\std{0.00010} & 0.98705\std{0.00010} & 0.98727\std{0.00002} \\
\midrule
\multicolumn{6}{l}{\textit{$1/\epsilon_B$ at $\epsilon_S=0.5$}} \\
L-GATr-slim\eqv{} & 546\std{10} & 539\std{6} & 513\std{11} & 535\std{20} & 542\std{21} \\
L-GATr-slim\eqv{} ens. & 577\std{8} & 572\std{10} & 523\std{12} & 550\std{7} & 575\std{7} \\
OmniLearned-S\pre{} & 565\std{6} & 536\std{12} & 501\std{6} & 561\std{9} & 558\std{18} \\
\midrule
\multicolumn{6}{l}{\textit{$1/\epsilon_B$ at $\epsilon_S=0.3$}} \\
L-GATr-slim\eqv{} & 2{,}374\std{103} & 2{,}261\std{38} & 2{,}171\std{116} & 2{,}242\std{149} & 2{,}274\std{141} \\
L-GATr-slim\eqv{} ens. & 2{,}529\std{134} & 2{,}313\std{30} & 2{,}263\std{52} & 2{,}331\std{63} & 2{,}351\std{55} \\
OmniLearned-S\pre{} & 2{,}325\std{121} & 2{,}483\std{120} & 2{,}045\std{121} & 2{,}318\std{88} & 2{,}345\std{177} \\
\bottomrule
\end{tabular}
}

\end{table}

\FloatBarrier
\section{Background rejection of every student}
\label{app:matrix_rej}

\begin{table}[h]
\centering
\caption{Background rejection $1/\epsilon_B$ at $\epsilon_S=0.5$ (top) and $\epsilon_S=0.3$ (bottom) of every student (columns), trained without a teacher and distilled from each teacher (rows). Means and standard deviations over three runs.}
\label{tab:matrix_rej}
\vspace{4pt}
{\small
\begin{tabular}{lrrrrr}
\toprule
& \multicolumn{5}{c}{Student} \\ \cmidrule(lr){2-6}
Teacher & Transformer & LorentzNet\eqv{} & ParT & MIParT & L-GATr-slim\eqv{} \\
\midrule
\multicolumn{6}{l}{\textit{$1/\epsilon_B$ at $\epsilon_S=0.5$}} \\
No teacher & 393\std{5} & 413\std{16} & 384\std{7} & 410\std{5} & 425\std{20} \\
ParT & 441\std{20} & 446\std{25} & 423\std{12} & 441\std{12} & 458\std{12} \\
MIParT & 482\std{9} & 488\std{2} & 448\std{10} & 474\std{11} & 487\std{9} \\
L-GATr\eqv{} & 537\std{14} & 527\std{9} & 512\std{8} & 523\std{9} & 533\std{24} \\
L-GATr-slim\eqv{} & 546\std{10} & 539\std{6} & 513\std{11} & 535\std{20} & 542\std{21} \\
OmniLearned-S\pre{} & 565\std{6} & 536\std{12} & 501\std{6} & \textbf{561}\std{9} & 558\std{18} \\
ParT-f.t.\pre{} & \textbf{604}\std{35} & \textbf{586}\std{5} & \textbf{525}\std{21} & 558\std{19} & \textbf{568}\std{8} \\
\midrule
\multicolumn{6}{l}{\textit{$1/\epsilon_B$ at $\epsilon_S=0.3$}} \\
No teacher & 1{,}612\std{121} & 1{,}714\std{83} & 1{,}603\std{76} & 1{,}767\std{83} & 1{,}704\std{80} \\
ParT & 1{,}826\std{143} & 1{,}897\std{96} & 1{,}733\std{69} & 1{,}857\std{73} & 1{,}937\std{72} \\
MIParT & 1{,}834\std{95} & 1{,}758\std{71} & 1{,}660\std{48} & 1{,}721\std{117} & 1{,}985\std{87} \\
L-GATr\eqv{} & 2{,}294\std{166} & 2{,}260\std{37} & 2{,}141\std{26} & 2{,}173\std{79} & 2{,}181\std{77} \\
L-GATr-slim\eqv{} & 2{,}374\std{103} & 2{,}261\std{38} & \textbf{2{,}171}\std{116} & 2{,}242\std{149} & 2{,}274\std{141} \\
OmniLearned-S\pre{} & 2{,}325\std{121} & \textbf{2{,}483}\std{120} & 2{,}045\std{121} & 2{,}318\std{88} & 2{,}345\std{177} \\
ParT-f.t.\pre{} & \textbf{2{,}493}\std{78} & 2{,}405\std{77} & 2{,}157\std{104} & \textbf{2{,}393}\std{172} & \textbf{2{,}672}\std{233} \\
\bottomrule
\end{tabular}
}

\end{table}

\FloatBarrier
\section{Scaling studies}
\label{app:results}

\begin{figure}[h]
\centering
\includegraphics[width=\linewidth]{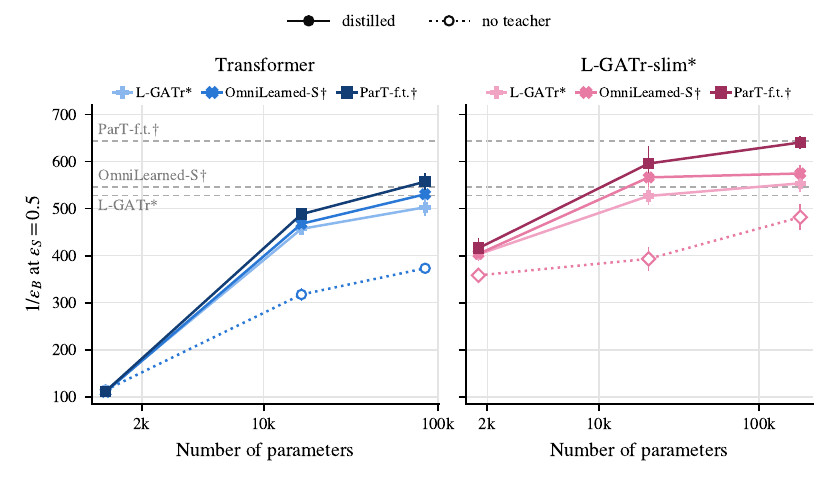}
\caption{Width study: background rejection $1/\epsilon_B$ at $\epsilon_S=0.5$ of ten-block students of decreasing width, trained without a teacher (dotted, hollow) and distilled from three teachers (solid, filled, darker for a stronger teacher). Error bars show the standard deviation over three runs, and grey dashed lines show the rejection of the teachers.}
\label{fig:width_rej50}
\end{figure}

\begin{figure}[h]
\centering
\includegraphics[width=\linewidth]{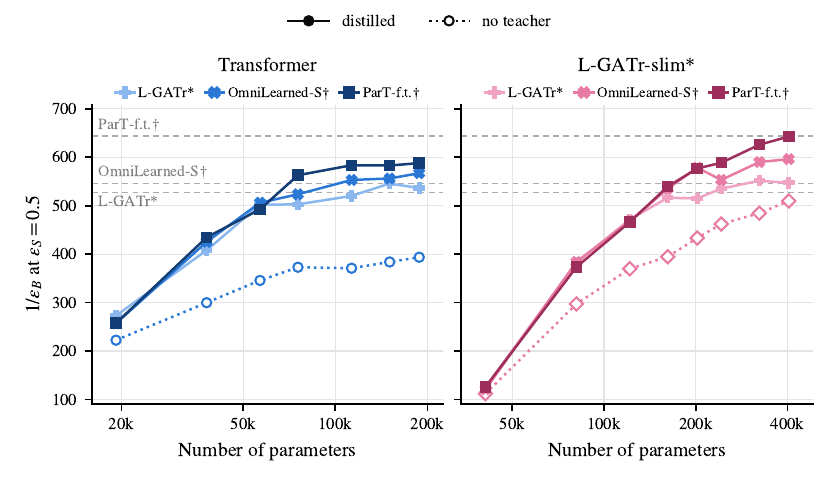}
\caption{Depth study: background rejection $1/\epsilon_B$ at $\epsilon_S=0.5$ of students of fixed width with a varying number of blocks, trained without a teacher (dotted, hollow) and distilled from three teachers (solid, filled, darker for a stronger teacher), one run each. Grey dashed lines show the rejection of the teachers.}
\label{fig:depth_rej50}
\end{figure}

Figures~\ref{fig:width_rej50} and~\ref{fig:depth_rej50} show the scaling studies at $\epsilon_S=0.5$, Table~\ref{tab:size_other} gives the width study with L-GATr and OmniLearned-S as teachers and Tables~\ref{tab:depth_main} and~\ref{tab:depth_other} give the depth study in full.

\begin{table}[h]
\centering
\caption{Width study: AUC and background rejection $1/\epsilon_B$ at $\epsilon_S=0.5$ and $0.3$ of ten-block students of three widths, distilled from L-GATr and OmniLearned-S. Means and standard deviations over three runs.}
\label{tab:size_other}
\vspace{4pt}
{\footnotesize
\setlength{\tabcolsep}{3.5pt}
\begin{tabular}{lrlrrlrr}
\toprule
& & \multicolumn{3}{c}{L-GATr\eqv{}} & \multicolumn{3}{c}{OmniLearned-S\pre{}} \\ \cmidrule(lr){3-5} \cmidrule(lr){6-8}
& & & \multicolumn{2}{c}{$1/\epsilon_B$} & & \multicolumn{2}{c}{$1/\epsilon_B$} \\ \cmidrule(lr){4-5} \cmidrule(lr){7-8}
Student & Params & AUC & $\epsilon_S{=}0.5$ & $\epsilon_S{=}0.3$ & AUC & $\epsilon_S{=}0.5$ & $\epsilon_S{=}0.3$ \\
\midrule
Transformer & 1.2k & 0.97391\std{0.00031} & 113\std{5} & 391\std{26} & 0.97360\std{0.00038} & 111\std{3} & 377\std{17} \\
 & 16k & 0.98590\std{0.00009} & 457\std{9} & 1{,}876\std{142} & 0.98615\std{0.00021} & 468\std{20} & 1{,}976\std{120} \\
 & 84k & 0.98658\std{0.00011} & 503\std{19} & 2{,}230\std{133} & 0.98669\std{0.00006} & 530\std{6} & 2{,}152\std{183} \\
\midrule
L-GATr-slim\eqv{} & 1.8k & 0.98560\std{0.00012} & 403\std{2} & 1{,}773\std{67} & 0.98543\std{0.00046} & 404\std{16} & 1{,}731\std{167} \\
 & 20k & 0.98683\std{0.00018} & 527\std{20} & 2{,}234\std{153} & 0.98719\std{0.00027} & 566\std{20} & 2{,}430\std{162} \\
 & 181k & 0.98705\std{0.00004} & 554\std{19} & 2{,}292\std{60} & 0.98758\std{0.00004} & 575\std{17} & 2{,}603\std{102} \\
\bottomrule
\end{tabular}
}

\end{table}

\begin{table}[h]
\centering
\caption{Depth study: AUC and background rejection $1/\epsilon_B$ at $\epsilon_S=0.5$ and $0.3$ of students of fixed width with a varying number of blocks, trained without a teacher and distilled from ParT-f.t., one run each.}
\label{tab:depth_main}
\vspace{4pt}
{\footnotesize
\setlength{\tabcolsep}{3.5pt}
\begin{tabular}{lrrlrrlrr}
\toprule
& & & \multicolumn{3}{c}{No teacher} & \multicolumn{3}{c}{ParT-f.t.\pre{}} \\ \cmidrule(lr){4-6} \cmidrule(lr){7-9}
& & & & \multicolumn{2}{c}{$1/\epsilon_B$} & & \multicolumn{2}{c}{$1/\epsilon_B$} \\ \cmidrule(lr){5-6} \cmidrule(lr){8-9}
Student & Blocks & Params & AUC & $\epsilon_S{=}0.5$ & $\epsilon_S{=}0.3$ & AUC & $\epsilon_S{=}0.5$ & $\epsilon_S{=}0.3$ \\
\midrule
Transformer & 0 & 433 & 0.86367 & 10 & 16 & 0.86364 & 10 & 16 \\
 & 1 & 19k & 0.98143 & 222 & 869 & 0.98275 & 257 & 1{,}128 \\
 & 2 & 38k & 0.98360 & 300 & 1{,}224 & 0.98577 & 435 & 1{,}771 \\
 & 3 & 57k & 0.98459 & 346 & 1{,}355 & 0.98648 & 492 & 2{,}103 \\
 & 4 & 76k & 0.98486 & 373 & 1{,}378 & 0.98681 & 563 & 2{,}294 \\
 & 6 & 113k & 0.98495 & 371 & 1{,}520 & 0.98709 & 584 & 2{,}125 \\
 & 8 & 151k & 0.98502 & 384 & 1{,}615 & 0.98723 & 583 & 2{,}433 \\
 & 10 & 188k & 0.98508 & 394 & 1{,}485 & 0.98725 & 588 & 2{,}524 \\
\midrule
L-GATr-slim\eqv{} & 1 & 41k & 0.97492 & 112 & 330 & 0.97621 & 126 & 410 \\
 & 2 & 81k & 0.98399 & 297 & 1{,}141 & 0.98497 & 373 & 1{,}496 \\
 & 3 & 122k & 0.98550 & 370 & 1{,}457 & 0.98643 & 466 & 1{,}887 \\
 & 4 & 162k & 0.98588 & 395 & 1{,}577 & 0.98709 & 540 & 2{,}404 \\
 & 5 & 202k & 0.98632 & 433 & 1{,}726 & 0.98738 & 577 & 2{,}619 \\
 & 6 & 243k & 0.98654 & 462 & 1{,}980 & 0.98751 & 589 & 2{,}766 \\
 & 8 & 323k & 0.98672 & 485 & 2{,}103 & 0.98765 & 626 & 2{,}692 \\
 & 10 & 404k & 0.98687 & 510 & 1{,}939 & 0.98771 & 643 & 2{,}969 \\
\bottomrule
\end{tabular}
}

\end{table}

\begin{table}[h]
\centering
\caption{Depth study: AUC and background rejection $1/\epsilon_B$ at $\epsilon_S=0.5$ and $0.3$ of students of fixed width with a varying number of blocks, distilled from L-GATr and OmniLearned-S, one run each.}
\label{tab:depth_other}
\vspace{4pt}
{\footnotesize
\setlength{\tabcolsep}{3.5pt}
\begin{tabular}{lrrlrrlrr}
\toprule
& & & \multicolumn{3}{c}{L-GATr\eqv{}} & \multicolumn{3}{c}{OmniLearned-S\pre{}} \\ \cmidrule(lr){4-6} \cmidrule(lr){7-9}
& & & & \multicolumn{2}{c}{$1/\epsilon_B$} & & \multicolumn{2}{c}{$1/\epsilon_B$} \\ \cmidrule(lr){5-6} \cmidrule(lr){8-9}
Student & Blocks & Params & AUC & $\epsilon_S{=}0.5$ & $\epsilon_S{=}0.3$ & AUC & $\epsilon_S{=}0.5$ & $\epsilon_S{=}0.3$ \\
\midrule
Transformer & 0 & 433 & 0.86357 & 10 & 16 & 0.86364 & 10 & 16 \\
 & 1 & 19k & 0.98295 & 273 & 1{,}091 & 0.98261 & 257 & 1{,}020 \\
 & 2 & 38k & 0.98549 & 408 & 1{,}836 & 0.98557 & 425 & 1{,}711 \\
 & 3 & 57k & 0.98641 & 502 & 2{,}148 & 0.98647 & 506 & 2{,}040 \\
 & 4 & 76k & 0.98645 & 503 & 1{,}923 & 0.98666 & 524 & 1{,}936 \\
 & 6 & 113k & 0.98673 & 520 & 2{,}103 & 0.98690 & 553 & 2{,}082 \\
 & 8 & 151k & 0.98672 & 546 & 2{,}219 & 0.98704 & 556 & 2{,}219 \\
 & 10 & 188k & 0.98688 & 536 & 2{,}375 & 0.98710 & 567 & 2{,}433 \\
\midrule
L-GATr-slim\eqv{} & 1 & 41k & 0.97619 & 125 & 409 & 0.97549 & 121 & 381 \\
 & 2 & 81k & 0.98516 & 382 & 1{,}442 & 0.98507 & 384 & 1{,}530 \\
 & 3 & 122k & 0.98639 & 472 & 1{,}923 & 0.98637 & 468 & 1{,}827 \\
 & 4 & 162k & 0.98678 & 516 & 2{,}348 & 0.98708 & 535 & 2{,}166 \\
 & 5 & 202k & 0.98693 & 515 & 2{,}269 & 0.98729 & 579 & 2{,}393 \\
 & 6 & 243k & 0.98698 & 535 & 2{,}321 & 0.98735 & 553 & 2{,}375 \\
 & 8 & 323k & 0.98708 & 552 & 2{,}414 & 0.98761 & 590 & 2{,}622 \\
 & 10 & 404k & 0.98709 & 547 & 2{,}348 & 0.98765 & 596 & 2{,}622 \\
\bottomrule
\end{tabular}
}

\end{table}

\FloatBarrier
\section{What every student learns from its teacher}
\label{app:learn}

\begin{figure}[h]
\centering
\includegraphics[width=\linewidth]{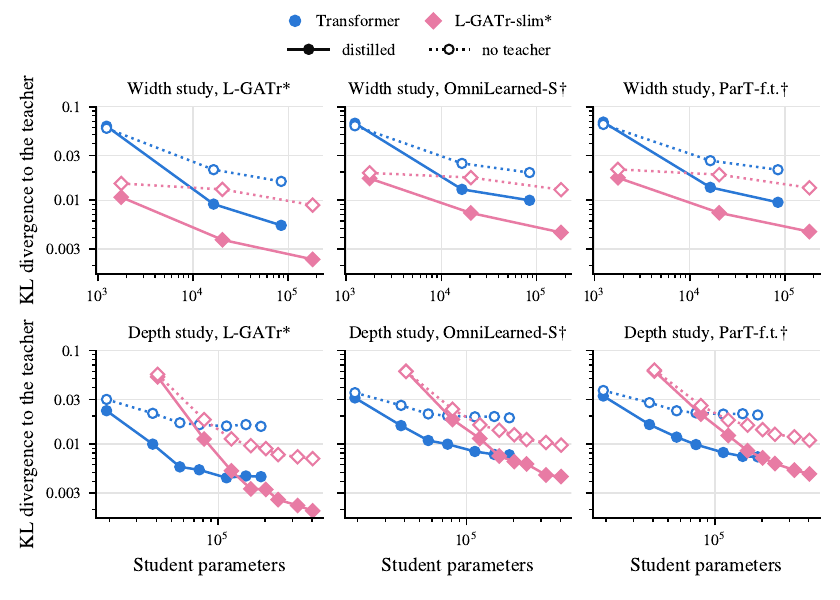}
\caption{KL divergence from the teacher's predicted probabilities to the student's against the number of student parameters, for the three teachers of the scaling studies, in the width study (top, means over three runs) and the depth study (bottom, one run, three at the depth of Section~\ref{sec:matrix}). Transformer and L-GATr-slim are shown distilled (solid, filled) and trained without a teacher (dotted, hollow).}
\label{fig:learn_size}
\end{figure}

\begin{figure}[h]
\centering
\includegraphics[width=\linewidth]{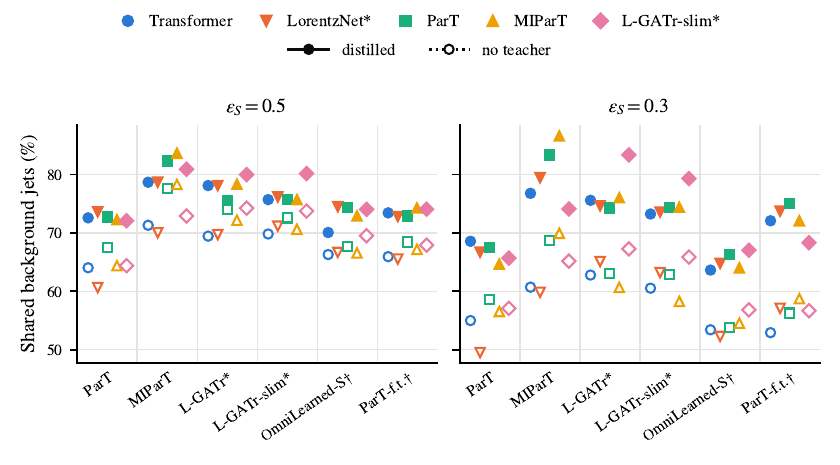}
\caption{Of the background jets each teacher accepts at $\epsilon_S=0.5$ (left) and $\epsilon_S=0.3$ (right), the percentage every student of Section~\ref{sec:matrix} also accepts at the same signal efficiency, distilled (filled) and trained without a teacher (hollow). Means over three runs.}
\label{fig:learn_shared}
\end{figure}

\begin{figure}[h]
\centering
\includegraphics[width=\linewidth]{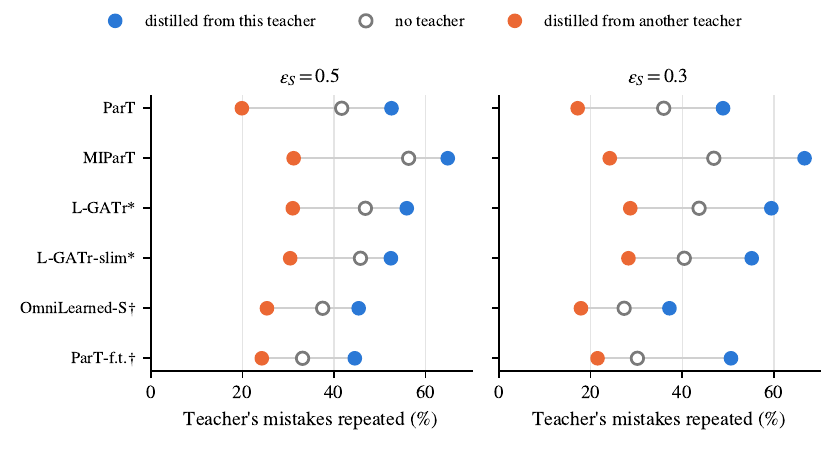}
\caption{For each teacher, the percentage of the background jets it accepts and another teacher rejects that the student also accepts at $\epsilon_S=0.5$ (left) and $\epsilon_S=0.3$ (right), when distilled from this teacher, distilled from the other teacher or trained without a teacher. Means over the five other teachers, the five students of Section~\ref{sec:matrix} and three runs.}
\label{fig:learn_distinctive}
\end{figure}

\begin{figure}[h]
\centering
\includegraphics[width=\linewidth]{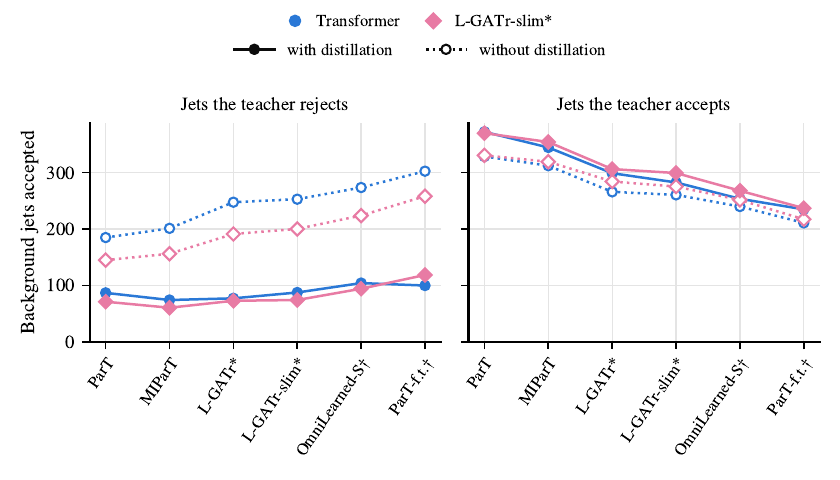}
\caption{Background jets Transformer and L-GATr-slim accept at $\epsilon_S=0.5$ among those each teacher rejects (left) and accepts (right), with distillation (solid, filled) and without (dotted, hollow). Means over three runs.}
\label{fig:learn_accepted}
\end{figure}

\begin{table}[h]
\centering
\caption{Background jets every student of Section~\ref{sec:matrix} accepts at $\epsilon_S=0.5$, without and with distillation, among the jets each teacher rejects (top) and accepts (bottom). Means over three runs.}
\label{tab:accepted}
\vspace{4pt}
{\footnotesize
\setlength{\tabcolsep}{3.4pt}
\begin{tabular}{lrrrrrrrrrr}
\toprule
 & \multicolumn{2}{c}{Transformer} & \multicolumn{2}{c}{LorentzNet\eqv{}} & \multicolumn{2}{c}{ParT} & \multicolumn{2}{c}{MIParT} & \multicolumn{2}{c}{L-GATr-slim\eqv{}} \\
\cmidrule(lr){2-3} \cmidrule(lr){4-5} \cmidrule(lr){6-7} \cmidrule(lr){8-9} \cmidrule(lr){10-11}
Teacher & without & with & without & with & without & with & without & with & without & with \\
\midrule
\multicolumn{11}{l}{\textit{Background jets the teacher rejects}} \\
ParT & 185 & 87 & 178 & 76 & 179 & 105 & 163 & 87 & 145 & 71 \\
MIParT & 201 & 74 & 183 & 69 & 186 & 91 & 150 & 60 & 156 & 61 \\
L-GATr\eqv{} & 248 & 77 & 222 & 85 & 243 & 105 & 217 & 87 & 191 & 73 \\
L-GATr-slim\eqv{} & 253 & 88 & 224 & 91 & 255 & 111 & 229 & 95 & 200 & 74 \\
OmniLearned-S\pre{} & 274 & 104 & 248 & 107 & 281 & 134 & 252 & 96 & 224 & 94 \\
ParT-f.t.\pre{} & 303 & 100 & 280 & 112 & 307 & 152 & 278 & 125 & 258 & 119 \\
\midrule
\multicolumn{11}{l}{\textit{Background jets the teacher accepts}} \\
ParT & 329 & 372 & 311 & 378 & 347 & 373 & 330 & 371 & 331 & 370 \\
MIParT & 312 & 345 & 307 & 345 & 340 & 361 & 343 & 366 & 319 & 354 \\
L-GATr\eqv{} & 266 & 299 & 267 & 299 & 283 & 289 & 276 & 300 & 284 & 306 \\
L-GATr-slim\eqv{} & 261 & 283 & 266 & 284 & 271 & 283 & 264 & 283 & 275 & 299 \\
OmniLearned-S\pre{} & 240 & 254 & 241 & 270 & 245 & 269 & 241 & 264 & 252 & 268 \\
ParT-f.t.\pre{} & 211 & 235 & 210 & 233 & 219 & 233 & 215 & 238 & 217 & 237 \\
\bottomrule
\end{tabular}
}

\end{table}

\begin{figure}[h]
\centering
\includegraphics[width=\linewidth]{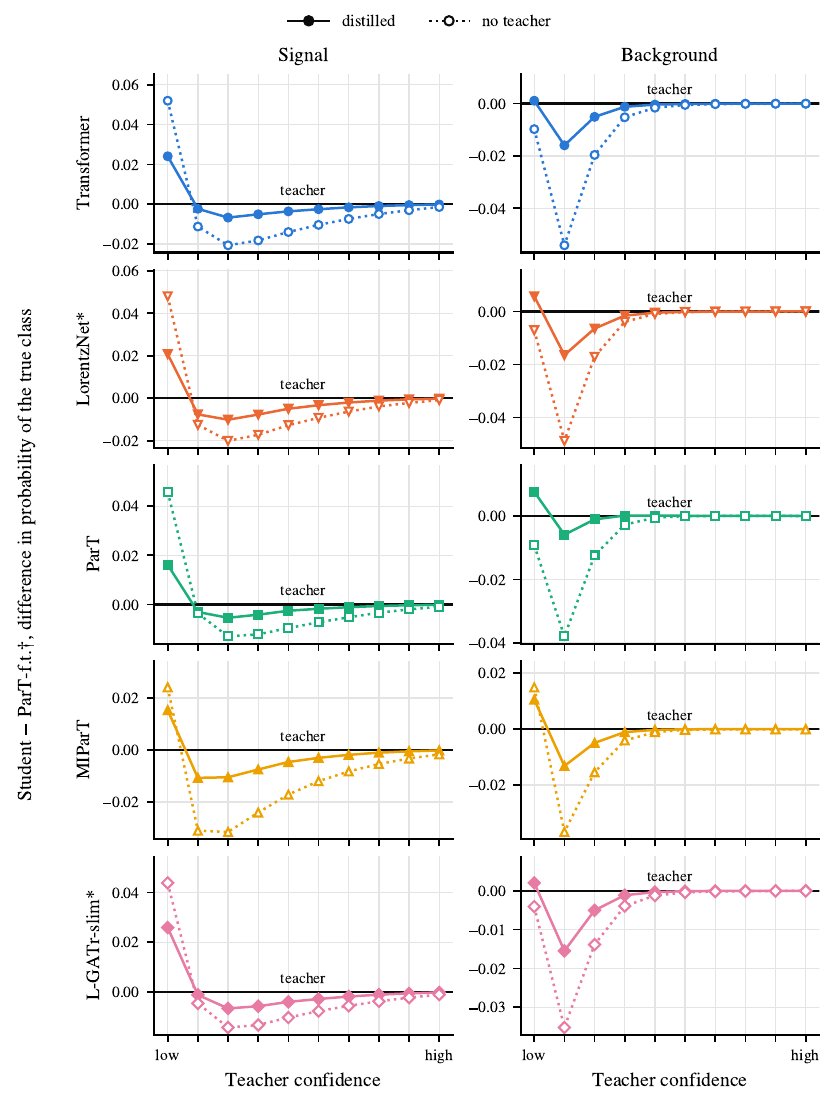}
\caption{Probability every student of Section~\ref{sec:matrix} assigns to the true class minus that assigned by ParT-f.t., for signal (left) and background (right) jets in ten groups of equal size and increasing \mbox{ParT-f.t.}\ confidence, distilled (solid, filled) and trained without a teacher (dotted, hollow). Means over three runs.}
\label{fig:learn_perjet}
\end{figure}

\begin{figure}[h]
\centering
\includegraphics[width=\linewidth]{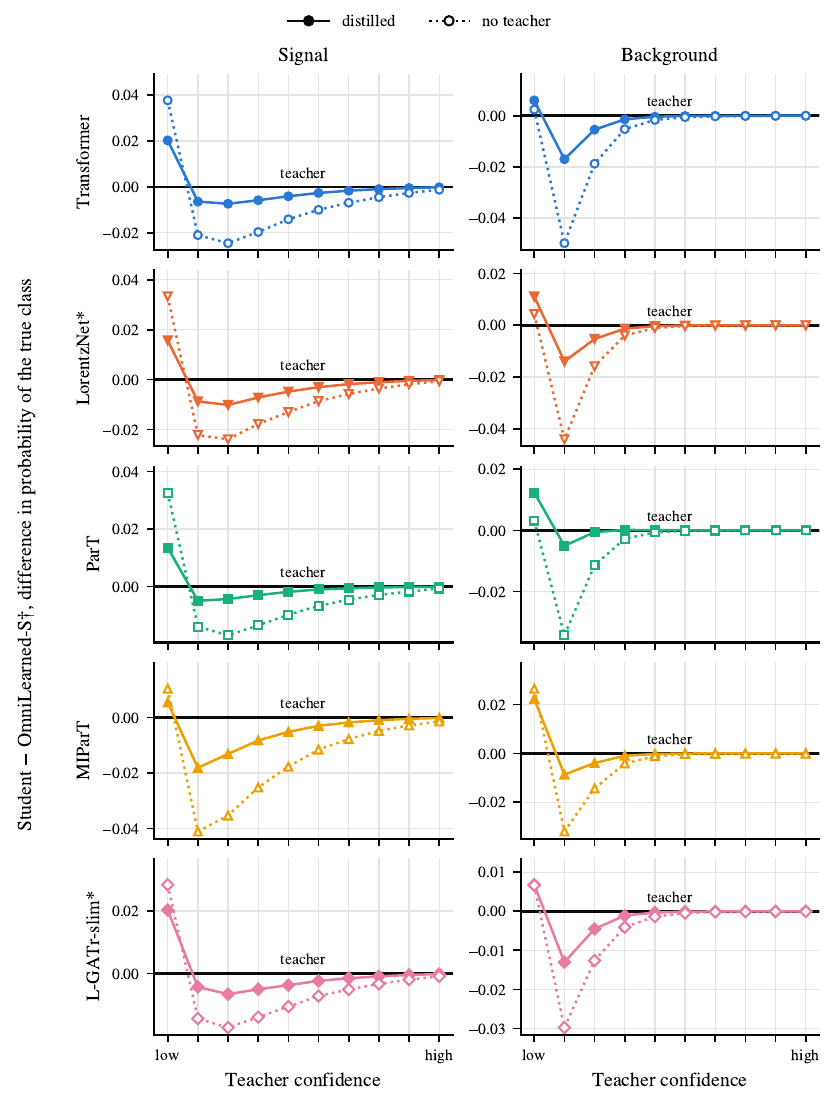}
\caption{Probability every student of Section~\ref{sec:matrix} assigns to the true class minus that assigned by OmniLearned-S, for signal (left) and background (right) jets in ten groups of equal size and increasing OmniLearned-S confidence, distilled (solid, filled) and trained without a teacher (dotted, hollow). Means over three runs.}
\label{fig:learn_perjet_omni}
\end{figure}

\begin{figure}[h]
\centering
\includegraphics[width=\linewidth]{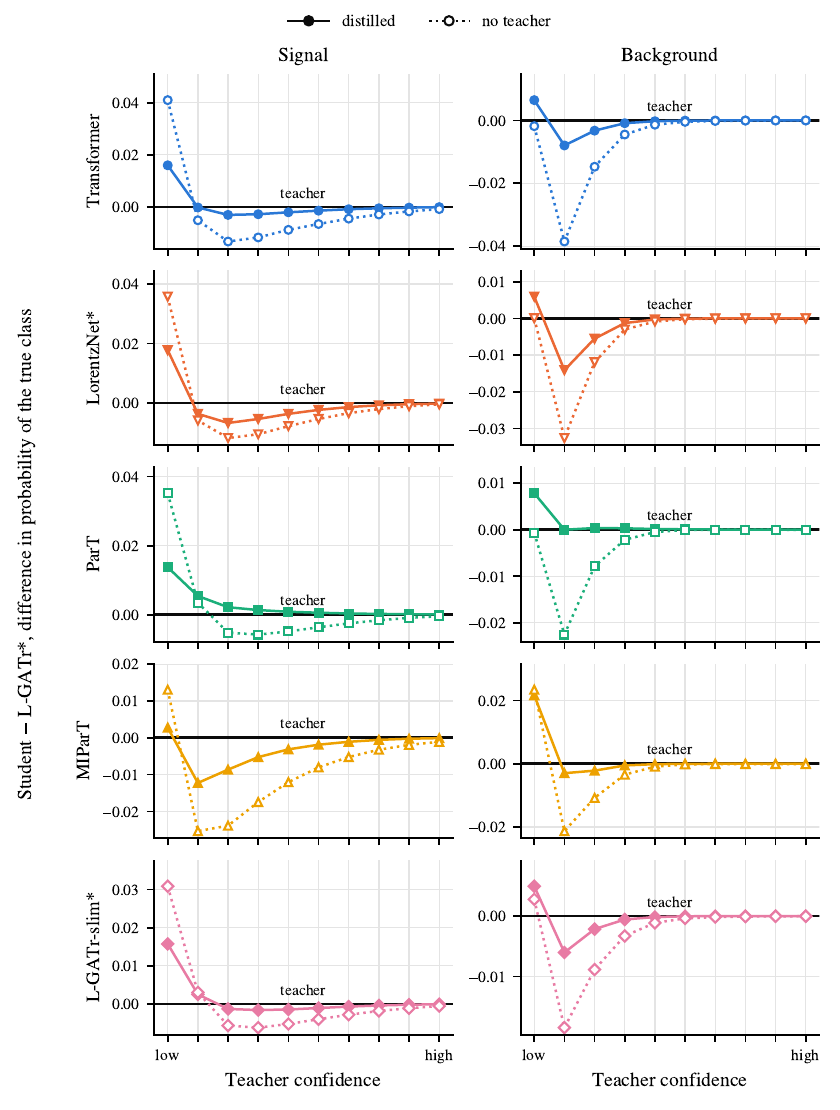}
\caption{Probability every student of Section~\ref{sec:matrix} assigns to the true class minus that assigned by L-GATr, for signal (left) and background (right) jets in ten groups of equal size and increasing L-GATr confidence, distilled (solid, filled) and trained without a teacher (dotted, hollow). Means over three runs.}
\label{fig:learn_perjet_lgatr}
\end{figure}

\FloatBarrier
\clearpage
\section{Inference benchmark}
\label{app:benchmark}

\begin{figure}[h]
\centering
\includegraphics[width=\linewidth]{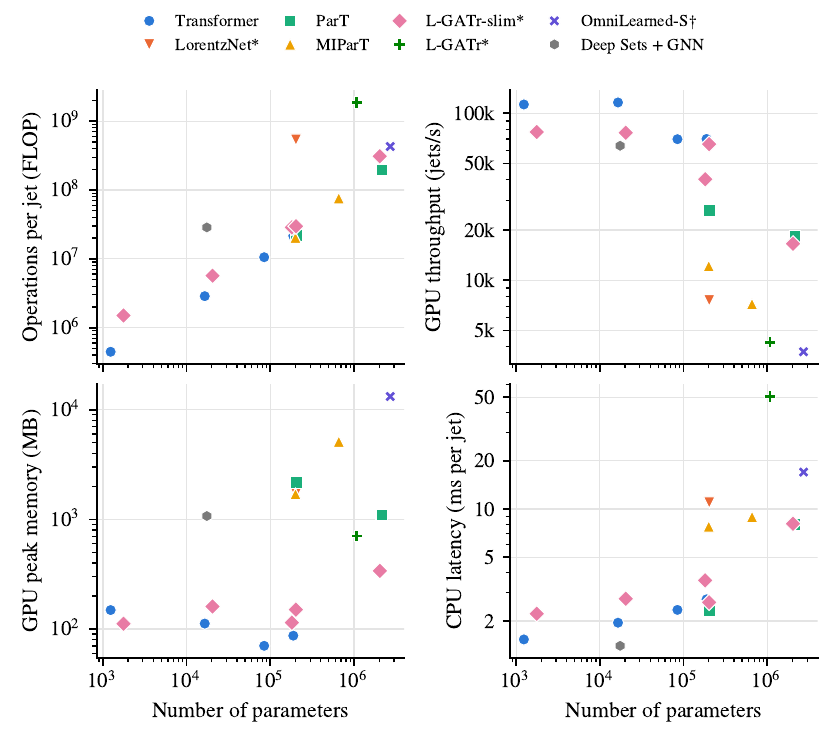}
\caption{Operations per jet, GPU throughput and peak memory at a batch of 512 jets and CPU latency for one jet on one core against the number of parameters, for every model of Table~\ref{tab:cost}, on real test jets in the fastest 32-bit setting.}
\label{fig:cost_params}
\end{figure}

\subsection{Execution settings}
We time every model in several 32-bit settings and report the fastest on each device (Table~\ref{tab:settings}).
Every model runs eager and compiled with \texttt{torch.compile}.
Models that accept both run with packed particle lists, which concatenate the particles of all jets in a batch, and with lists padded to the longest jet in the batch.
Transformer at every size runs with xFormers attention, L-GATr and L-GATr-slim with xFormers and FlexAttention, and the other models with their built-in attention.
On the GPU every setting runs once, and the fastest runs three more times.
On the CPU every setting runs three times on one fixed core, and the fastest setting is the one with the lowest median.
The smallest models are fastest with padded lists, which hold more memory than packed ones, so some of them reach a higher peak memory than larger models in Table~\ref{tab:cost}.

\begin{table}[h]
\centering
\caption{Execution settings timed for every model: particle lists and attention kernels (Flex: FlexAttention), each eager and compiled, and the fastest setting on each device, which Tables~\ref{tab:cost} and~\ref{tab:cost_rnd50} report.}
\label{tab:settings}
\vspace{4pt}
\setlength{\tabcolsep}{4pt}
{\footnotesize
\begin{tabular}{lllll}
\toprule
 & & & \multicolumn{2}{c}{Fastest setting} \\ \cmidrule(lr){4-5}
Model & Particle lists & Attention & GPU & CPU \\
\midrule
ParT & padded & built-in & padded, compiled, built-in & padded, compiled, built-in \\
MIParT & padded & built-in & padded, compiled, built-in & padded, compiled, built-in \\
L-GATr\eqv{} & packed, padded & xFormers, Flex & packed, compiled, xFormers & padded, compiled, xFormers \\
L-GATr-slim\eqv{} & packed, padded & xFormers, Flex & packed, compiled, xFormers & packed, compiled, xFormers \\
OmniLearned-S\pre{} & padded & built-in & padded, compiled, built-in & padded, compiled, built-in \\
ParT-f.t.\pre{} & padded & built-in & padded, compiled, built-in & padded, compiled, built-in \\
Transformer 188k & packed, padded & xFormers & packed, eager, xFormers & packed, compiled, xFormers \\
LorentzNet\eqv{} 203k & packed & built-in & packed, compiled, built-in & packed, compiled, built-in \\
ParT 205k & padded & built-in & padded, compiled, built-in & padded, compiled, built-in \\
MIParT 200k & padded & built-in & padded, compiled, built-in & padded, compiled, built-in \\
L-GATr-slim\eqv{} 202k & packed, padded & xFormers, Flex & packed, compiled, xFormers & padded, compiled, xFormers \\
Transformer 84k & packed, padded & xFormers & packed, eager, xFormers & packed, compiled, xFormers \\
Transformer 16k & packed, padded & xFormers & padded, compiled, xFormers & packed, compiled, xFormers \\
Transformer 1.2k & packed, padded & xFormers & padded, compiled, xFormers & padded, compiled, xFormers \\
L-GATr-slim\eqv{} 181k & packed, padded & xFormers, Flex & packed, compiled, xFormers & packed, compiled, xFormers \\
L-GATr-slim\eqv{} 20k & packed, padded & xFormers, Flex & padded, compiled, xFormers & packed, compiled, xFormers \\
L-GATr-slim\eqv{} 1.8k & packed, padded & xFormers, Flex & padded, compiled, xFormers & padded, compiled, xFormers \\
Deep Sets + GNN & padded & built-in & padded, compiled, built-in & padded, compiled, built-in \\
\bottomrule
\end{tabular}
}

\end{table}

\subsection{Precision}
\label{app:precision}

We run every model in full 32-bit precision, as in training.
An earlier cost benchmark of these models~\citep{favaro2026virtues} uses TF32 matrix multiplication, which keeps only 10 of the 23 mantissa bits of a 32-bit number.
Under TF32, Transformer gives the same outputs, but L-GATr and L-GATr-slim lose performance, the small L-GATr-slim students most strongly.
These networks act on Minkowski products of four-momenta, which for nearly collinear particles are small differences of large numbers, so rounding the momenta changes them strongly.
Transformer takes features of order one, such as logarithms of momenta and angular distances, and is unaffected.

\subsection{Jets with 50 particles}
\label{app:rnd50}

The earlier cost benchmark~\citep{favaro2026virtues} measures CPU latency on jets with exactly 50 particles and random momenta.
Table~\ref{tab:cost_rnd50} repeats our measurements in this setting, in which the cost no longer depends on how many particles the test jets carry.
On the GPU each setting runs once.

\begin{table}[h]
\centering
\caption{Inference cost on jets with exactly 50 particles and random momenta: GPU throughput and peak memory at a batch of 512 jets and CPU latency for one jet on one core, each in the fastest 32-bit setting.}
\label{tab:cost_rnd50}
\vspace{4pt}
{\small
\begin{tabular}{lrrrr}
\toprule
 & & \multicolumn{2}{c}{GPU} & CPU \\ \cmidrule(lr){3-4}
Model & Params & jets/s & MB & ms/jet \\
\midrule
\multicolumn{5}{l}{\textit{Teachers}} \\
ParT & 2.1M & 38{,}503 & 388 & 7.3 \\
MIParT & 0.7M & 28{,}320 & 1{,}007 & 8.5 \\
L-GATr\eqv{} & 1.1M & 5{,}641 & 785 & 47.6 \\
L-GATr-slim\eqv{} & 2.0M & 17{,}485 & 351 & 7.6 \\
OmniLearned-S\pre{} & 2.7M & 10{,}550 & 2{,}602 & 14.9 \\
ParT-f.t.\pre{} & 2.1M & 38{,}509 & 388 & 7.3 \\
\midrule
\multicolumn{5}{l}{\textit{Students}} \\
Transformer & 188k & 93{,}179 & 82 & 2.7 \\
LorentzNet\eqv{} & 203k & 7{,}858 & 1{,}589 & 9.7 \\
ParT & 205k & 82{,}427 & 379 & 2.2 \\
MIParT & 200k & 41{,}164 & 353 & 7.5 \\
L-GATr-slim\eqv{} & 202k & 84{,}660 & 145 & 2.5 \\
\midrule
\multicolumn{5}{l}{\textit{Width study}} \\
Transformer & 84k & 121{,}640 & 62 & 2.3 \\
Transformer & 16k & 137{,}684 & 40 & 1.9 \\
Transformer & 1.2k & 130{,}787 & 43 & 1.5 \\
L-GATr-slim\eqv{} & 181k & 71{,}114 & 118 & 3.3 \\
L-GATr-slim\eqv{} & 20k & 87{,}116 & 63 & 2.7 \\
L-GATr-slim\eqv{} & 1.8k & 75{,}042 & 46 & 2.2 \\
\midrule
\multicolumn{5}{l}{\textit{Distilled from OmniLearned-L\pre{}}} \\
Deep Sets + GNN & 17k & 73{,}943 & 659 & 1.2 \\
\bottomrule
\end{tabular}
}

\end{table}

\FloatBarrier
\section{Performance and inference cost with every teacher}
\label{app:cost_full}

A student costs the same with every teacher, and only its performance changes.
Tables~\ref{tab:cost_scratch} to~\ref{tab:cost_omnilearned_s} give the performance and cost of the students trained without a teacher and of those distilled from each teacher other than ParT-f.t., together with that teacher.

\begin{table}[h]
\centering
\caption{Performance and inference cost of the students trained without a teacher.}
\label{tab:cost_scratch}
\vspace{4pt}
{\small\setlength{\tabcolsep}{4pt}
\begin{tabular}{lrrrrrrrr}
\toprule
 & & & \multicolumn{2}{c}{$1/\epsilon_B$} & & \multicolumn{2}{c}{GPU} & CPU \\ \cmidrule(lr){4-5} \cmidrule(lr){7-8}
Model & Params & AUC & $\epsilon_S{=}0.5$ & $\epsilon_S{=}0.3$ & MFLOP & jets/s & MB & ms/jet \\
\midrule
\multicolumn{9}{l}{\textit{Students}} \\
Transformer & 188k & 0.98512 & 393\std{5} & 1{,}612\std{121} & 21.6 & 70{,}175 & 87 & 2.7 \\
LorentzNet\eqv{} & 203k & 0.98556 & 413\std{16} & 1{,}714\std{83} & 537 & 7{,}601 & 1{,}722 & 11.0 \\
ParT & 205k & 0.98565 & 384\std{7} & 1{,}603\std{76} & 21.9 & 26{,}228 & 2{,}166 & 2.3 \\
MIParT & 200k & 0.98580 & 410\std{5} & 1{,}767\std{83} & 20.4 & 12{,}181 & 1{,}709 & 7.8 \\
L-GATr-slim\eqv{} & 202k & 0.98622 & 425\std{20} & 1{,}704\std{80} & 29.7 & 65{,}308 & 150 & 2.6 \\
\midrule
\multicolumn{9}{l}{\textit{Width study}} \\
Transformer & 84k & 0.98495 & 373\std{8} & 1{,}431\std{46} & 10.6 & 69{,}945 & 70 & 2.3 \\
Transformer & 16k & 0.98388 & 318\std{14} & 1{,}246\std{57} & 2.87 & 115{,}996 & 112 & 2.0 \\
Transformer & 1.2k & 0.97422 & 115\std{4} & 398\std{27} & 0.45 & 112{,}829 & 149 & 1.5 \\
L-GATr-slim\eqv{} & 181k & 0.98635 & 482\std{28} & 1{,}900\std{178} & 28.6 & 40{,}269 & 115 & 3.6 \\
L-GATr-slim\eqv{} & 20k & 0.98539 & 393\std{26} & 1{,}579\std{122} & 5.70 & 76{,}488 & 161 & 2.8 \\
L-GATr-slim\eqv{} & 1.8k & 0.98491 & 358\std{5} & 1{,}449\std{86} & 1.50 & 77{,}262 & 111 & 2.2 \\
\bottomrule
\end{tabular}
}

\end{table}

\begin{table}[h]
\centering
\caption{Performance and inference cost of the students distilled from ParT and of ParT itself.}
\label{tab:cost_part}
\vspace{4pt}
{\small\setlength{\tabcolsep}{4pt}
\begin{tabular}{lrrrrrrrr}
\toprule
 & & & \multicolumn{2}{c}{$1/\epsilon_B$} & & \multicolumn{2}{c}{GPU} & CPU \\ \cmidrule(lr){4-5} \cmidrule(lr){7-8}
Model & Params & AUC & $\epsilon_S{=}0.5$ & $\epsilon_S{=}0.3$ & MFLOP & jets/s & MB & ms/jet \\
\midrule
\multicolumn{9}{l}{\textit{Teacher}} \\
ParT & 2.1M & 0.98571 & 394\std{14} & 1{,}586\std{19} & 197 & 18{,}198 & 1{,}099 & 7.9 \\
\midrule
\multicolumn{9}{l}{\textit{Students}} \\
Transformer & 188k & 0.98617 & 441\std{20} & 1{,}826\std{143} & 21.6 & 70{,}175 & 87 & 2.7 \\
LorentzNet\eqv{} & 203k & 0.98630 & 446\std{25} & 1{,}897\std{96} & 537 & 7{,}601 & 1{,}722 & 11.0 \\
ParT & 205k & 0.98607 & 423\std{12} & 1{,}733\std{69} & 21.9 & 26{,}228 & 2{,}166 & 2.3 \\
MIParT & 200k & 0.98624 & 441\std{12} & 1{,}857\std{73} & 20.4 & 12{,}181 & 1{,}709 & 7.8 \\
L-GATr-slim\eqv{} & 202k & 0.98643 & 458\std{12} & 1{,}937\std{72} & 29.7 & 65{,}308 & 150 & 2.6 \\
\bottomrule
\end{tabular}
}

\end{table}

\begin{table}[h]
\centering
\caption{Performance and inference cost of the students distilled from MIParT and of MIParT itself.}
\label{tab:cost_mipart}
\vspace{4pt}
{\small\setlength{\tabcolsep}{4pt}
\begin{tabular}{lrrrrrrrr}
\toprule
 & & & \multicolumn{2}{c}{$1/\epsilon_B$} & & \multicolumn{2}{c}{GPU} & CPU \\ \cmidrule(lr){4-5} \cmidrule(lr){7-8}
Model & Params & AUC & $\epsilon_S{=}0.5$ & $\epsilon_S{=}0.3$ & MFLOP & jets/s & MB & ms/jet \\
\midrule
\multicolumn{9}{l}{\textit{Teacher}} \\
MIParT & 0.7M & 0.98632 & 467\std{5} & 1{,}871\std{69} & 76.0 & 7{,}225 & 5{,}132 & 8.9 \\
\midrule
\multicolumn{9}{l}{\textit{Students}} \\
Transformer & 188k & 0.98646 & 482\std{9} & 1{,}834\std{95} & 21.6 & 70{,}175 & 87 & 2.7 \\
LorentzNet\eqv{} & 203k & 0.98655 & 488\std{2} & 1{,}758\std{71} & 537 & 7{,}601 & 1{,}722 & 11.0 \\
ParT & 205k & 0.98622 & 448\std{10} & 1{,}660\std{48} & 21.9 & 26{,}228 & 2{,}166 & 2.3 \\
MIParT & 200k & 0.98631 & 474\std{11} & 1{,}721\std{117} & 20.4 & 12{,}181 & 1{,}709 & 7.8 \\
L-GATr-slim\eqv{} & 202k & 0.98660 & 487\std{9} & 1{,}985\std{87} & 29.7 & 65{,}308 & 150 & 2.6 \\
\bottomrule
\end{tabular}
}

\end{table}

\begin{table}[h]
\centering
\caption{Performance and inference cost of the students distilled from L-GATr and of L-GATr itself.}
\label{tab:cost_lgatr2}
\vspace{4pt}
{\small\setlength{\tabcolsep}{4pt}
\begin{tabular}{lrrrrrrrr}
\toprule
 & & & \multicolumn{2}{c}{$1/\epsilon_B$} & & \multicolumn{2}{c}{GPU} & CPU \\ \cmidrule(lr){4-5} \cmidrule(lr){7-8}
Model & Params & AUC & $\epsilon_S{=}0.5$ & $\epsilon_S{=}0.3$ & MFLOP & jets/s & MB & ms/jet \\
\midrule
\multicolumn{9}{l}{\textit{Teacher}} \\
L-GATr\eqv{} & 1.1M & 0.98680 & 528\std{14} & 2{,}189\std{176} & 1{,}868 & 4{,}263 & 707 & 50.5 \\
\midrule
\multicolumn{9}{l}{\textit{Students}} \\
Transformer & 188k & 0.98680 & 537\std{14} & 2{,}294\std{166} & 21.6 & 70{,}175 & 87 & 2.7 \\
LorentzNet\eqv{} & 203k & 0.98671 & 527\std{9} & 2{,}260\std{37} & 537 & 7{,}601 & 1{,}722 & 11.0 \\
ParT & 205k & 0.98659 & 512\std{8} & 2{,}141\std{26} & 21.9 & 26{,}228 & 2{,}166 & 2.3 \\
MIParT & 200k & 0.98675 & 523\std{9} & 2{,}173\std{79} & 20.4 & 12{,}181 & 1{,}709 & 7.8 \\
L-GATr-slim\eqv{} & 202k & 0.98693 & 533\std{24} & 2{,}181\std{77} & 29.7 & 65{,}308 & 150 & 2.6 \\
\midrule
\multicolumn{9}{l}{\textit{Width study}} \\
Transformer & 84k & 0.98658 & 503\std{19} & 2{,}230\std{133} & 10.6 & 69{,}945 & 70 & 2.3 \\
Transformer & 16k & 0.98590 & 457\std{9} & 1{,}876\std{142} & 2.87 & 115{,}996 & 112 & 2.0 \\
Transformer & 1.2k & 0.97391 & 113\std{5} & 391\std{26} & 0.45 & 112{,}829 & 149 & 1.5 \\
L-GATr-slim\eqv{} & 181k & 0.98705 & 554\std{19} & 2{,}292\std{60} & 28.6 & 40{,}269 & 115 & 3.6 \\
L-GATr-slim\eqv{} & 20k & 0.98683 & 527\std{20} & 2{,}234\std{153} & 5.70 & 76{,}488 & 161 & 2.8 \\
L-GATr-slim\eqv{} & 1.8k & 0.98560 & 403\std{2} & 1{,}773\std{67} & 1.50 & 77{,}262 & 111 & 2.2 \\
\bottomrule
\end{tabular}
}

\end{table}

\begin{table}[h]
\centering
\caption{Performance and inference cost of the students distilled from L-GATr-slim and of L-GATr-slim itself.}
\label{tab:cost_lgatr2_slim}
\vspace{4pt}
{\small\setlength{\tabcolsep}{4pt}
\begin{tabular}{lrrrrrrrr}
\toprule
 & & & \multicolumn{2}{c}{$1/\epsilon_B$} & & \multicolumn{2}{c}{GPU} & CPU \\ \cmidrule(lr){4-5} \cmidrule(lr){7-8}
Model & Params & AUC & $\epsilon_S{=}0.5$ & $\epsilon_S{=}0.3$ & MFLOP & jets/s & MB & ms/jet \\
\midrule
\multicolumn{9}{l}{\textit{Teacher}} \\
L-GATr-slim\eqv{} & 2.0M & 0.98711 & 541\std{5} & 2{,}321\std{46} & 309 & 16{,}546 & 339 & 8.1 \\
\midrule
\multicolumn{9}{l}{\textit{Students}} \\
Transformer & 188k & 0.98691 & 546\std{10} & 2{,}374\std{103} & 21.6 & 70{,}175 & 87 & 2.7 \\
LorentzNet\eqv{} & 203k & 0.98685 & 539\std{6} & 2{,}261\std{38} & 537 & 7{,}601 & 1{,}722 & 11.0 \\
ParT & 205k & 0.98671 & 513\std{11} & 2{,}171\std{116} & 21.9 & 26{,}228 & 2{,}166 & 2.3 \\
MIParT & 200k & 0.98690 & 535\std{20} & 2{,}242\std{149} & 20.4 & 12{,}181 & 1{,}709 & 7.8 \\
L-GATr-slim\eqv{} & 202k & 0.98712 & 542\std{21} & 2{,}274\std{141} & 29.7 & 65{,}308 & 150 & 2.6 \\
\bottomrule
\end{tabular}
}

\end{table}

\begin{table}[h]
\centering
\caption{Performance and inference cost of the students distilled from OmniLearned-S and of OmniLearned-S itself.}
\label{tab:cost_omnilearned_s}
\vspace{4pt}
{\small\setlength{\tabcolsep}{4pt}
\begin{tabular}{lrrrrrrrr}
\toprule
 & & & \multicolumn{2}{c}{$1/\epsilon_B$} & & \multicolumn{2}{c}{GPU} & CPU \\ \cmidrule(lr){4-5} \cmidrule(lr){7-8}
Model & Params & AUC & $\epsilon_S{=}0.5$ & $\epsilon_S{=}0.3$ & MFLOP & jets/s & MB & ms/jet \\
\midrule
\multicolumn{9}{l}{\textit{Teacher}} \\
OmniLearned-S\pre{} & 2.7M & 0.98746 & 546\std{11} & 2{,}403\std{100} & 427 & 3{,}736 & 13{,}221 & 17.0 \\
\midrule
\multicolumn{9}{l}{\textit{Students}} \\
Transformer & 188k & 0.98705 & 565\std{6} & 2{,}325\std{121} & 21.6 & 70{,}175 & 87 & 2.7 \\
LorentzNet\eqv{} & 203k & 0.98713 & 536\std{12} & 2{,}483\std{120} & 537 & 7{,}601 & 1{,}722 & 11.0 \\
ParT & 205k & 0.98677 & 501\std{6} & 2{,}045\std{121} & 21.9 & 26{,}228 & 2{,}166 & 2.3 \\
MIParT & 200k & 0.98705 & 561\std{9} & 2{,}318\std{88} & 20.4 & 12{,}181 & 1{,}709 & 7.8 \\
L-GATr-slim\eqv{} & 202k & 0.98727 & 558\std{18} & 2{,}345\std{177} & 29.7 & 65{,}308 & 150 & 2.6 \\
\midrule
\multicolumn{9}{l}{\textit{Width study}} \\
Transformer & 84k & 0.98669 & 530\std{6} & 2{,}152\std{183} & 10.6 & 69{,}945 & 70 & 2.3 \\
Transformer & 16k & 0.98615 & 468\std{20} & 1{,}976\std{120} & 2.87 & 115{,}996 & 112 & 2.0 \\
Transformer & 1.2k & 0.97360 & 111\std{3} & 377\std{17} & 0.45 & 112{,}829 & 149 & 1.5 \\
L-GATr-slim\eqv{} & 181k & 0.98758 & 575\std{17} & 2{,}603\std{102} & 28.6 & 40{,}269 & 115 & 3.6 \\
L-GATr-slim\eqv{} & 20k & 0.98719 & 566\std{20} & 2{,}430\std{162} & 5.70 & 76{,}488 & 161 & 2.8 \\
L-GATr-slim\eqv{} & 1.8k & 0.98543 & 404\std{16} & 1{,}731\std{167} & 1.50 & 77{,}262 & 111 & 2.2 \\
\bottomrule
\end{tabular}
}

\end{table}

\end{document}